\documentclass[journal]{IEEEtran}
\usepackage{amsmath,cite}
\usepackage{amssymb}
\usepackage{epsfig}
\usepackage[ruled,vlined,linesnumbered]{algorithm2e}
\usepackage{algorithmic}
\usepackage{subfigure}
\usepackage{booktabs,makecell}
\usepackage{multirow}
\usepackage{threeparttable}

\ifCLASSINFOpdf
\else
\fi
\begin{document}
%
\title{Broad Learning System Based on Maximum Correntropy Criterion}
%
%
%

\author{Yunfei~Zheng,~\IEEEmembership{}
        Badong~Chen,~\IEEEmembership{Senior Member,~IEEE,}
        Shiyuan~Wang,~\IEEEmembership{Senior Member,~IEEE,}
        and Weiqun Wang,~\IEEEmembership{Member,~IEEE}
\thanks{Yunfei Zheng and Badong Chen (corresponding author) are with the
Institute of Artificial Intelligence and Robotics, Xi'an Jiaotong University,
Xi'an 710049, China (e-mail: zhengyf@stu.xjtu.edu.cn; chenbd@mail.xjtu.edu.cn).}
\thanks{Shiyuan~Wang is with the College of Electronic and Information Engineering, Southwest University, Chongqing 400715, China (e-mail: wsy@swu.edu.cn).}
\thanks{Weiqun Wang is with the State Key Laboratory
of Management and Control for Complex Systems, Institute
of Automation, Chinese Academy of Sciences, Beijing 100190, China
(e-mail: weiqun.wang@ia.ac.cn).}
\thanks{This work was supported by National Natural Science Foundation of
China (91648208, 61976175), National Natural Science Foundation-Shenzhen Joint
Research Program (U1613219), and The Key Project of Natural Science Basic Research Plan in Shaanxi Province of China (2019JZ-05).}
}

\maketitle

\begin{abstract}
As an effective and efficient discriminative learning method, Broad Learning System (BLS) has received increasing attention due to its outstanding performance in various regression and classification problems. However, the standard BLS is derived under the minimum mean square error (MMSE) criterion, which is, of course, not always a good choice due to its sensitivity to outliers. To enhance the robustness of BLS, we propose in this work to adopt the maximum correntropy criterion (MCC) to train the output weights, obtaining a correntropy based broad learning system (C-BLS). Thanks to the inherent superiorities of MCC, the proposed C-BLS is expected to achieve excellent robustness to
outliers while maintaining the original performance of the standard BLS in Gaussian or noise-free environment. In addition, three alternative incremental learning algorithms, derived from a weighted regularized least-squares solution rather than pseudoinverse formula, for C-BLS are developed.
With the incremental learning algorithms, the system can be updated quickly without the entire retraining process from the beginning, when some new samples arrive or the network deems to be expanded. Experiments on
various regression and classification datasets are reported to demonstrate the desirable performance of the
new methods.

\end{abstract}

%
\IEEEpeerreviewmaketitle

%
%
%
\section{Introduction}

Broad Learning System (BLS) \cite{BLS2018} is an emerging discriminative learning method, which has been shown with the potential to outperform some deep neural network based learning methods, such as multilayer perceptron-based methods (MLP) \cite{MLP}, deep belief networks (DBN) \cite{DBN}, and stacked auto encoders (SAE) \cite{SAE}.
To design a BLS, there are several necessary steps, including: 1) transforming the input data into general mapped features by some feature
mappings; 2) the generated mapping features are connected by nonlinear activation functions to form the so called the ``enhancement nodes"; 3) the mapped features and the ``enhancement nodes" are sent together into the output layer, and the corresponding output weights are obtained by the means of pseudo-inverse.
Since all weights and biases of the hidden layer units in BLS can be randomly generated and remain unchanged after that, we only need to train the weights between the hidden layer and the output layer, which brings great convenience to the training process. In addition, if some new samples arrive or the network deems to be expanded, several practical incremental learning algorithms were developed to guarantee
that the system can be remodeled quickly without the entire
retraining process from the beginning \cite{BLS2018}.
Thanks to these attractive features, BLS has received increasing attention \cite{BLS_U, BLS_F, BLS_SM, BLS_G, BLS_L1, BLS_R, BLS_REMOTE, BLS_ACCESS, BLS_ENHANCE, BLS_EEG, BLS_Multiview, BLS_facial, BLS_CNN} and been successfully applied in image recognition, face recognition, time series prediction, etc.

The standard BLS, however, takes the minimum mean square error (MMSE) criterion as a default choice of the optimization criterion in training the network output weights.
Although MMSE criterion is computationally efficient and can provide good performance in Gaussian or noise-free environments, it will degrade the performance of BLS in complicated noise environments especially when data are contaminated by some outliers.
To address this issue, several alternative optimization criteria which combine $l_{1}$-norm with different regularization terms were proposed to train the output weights of BLS, generating a class of robust BLS (RBLS) variants \cite{BLS_L1}. Since $l_{1}$-norm is less sensitive to outliers, the robustness of BLS has been significantly improved. Along the same line, Chu et al. \cite{WBLS} put forward the weighted BLS (WBLS). With the well-designed weighted penalty factor, WBLS has shown good robustness in nonlinear industrial process.
Another representative work to improve the robustness of BLS is the robust manifold BLS (RM-BLS) \cite{RMBLS}. By introducing the manifold embedding and random perturbation approximation, the robust mapping features can be expected in some special application scenarios, like the noisy chaotic time series prediction.
Therefore, RM-BLS also has the ability to improve the robustness of BLS.

Although the aforementioned robust BLS variants can be good candidates when some training data are disturbed by outliers, they suffer from some drawbacks.
For example, due to computational complexity, the incremental learning algorithms have not been provided under $l_{1}$-norm based optimization criteria, even though they are one of the most important features of the standard BLS.
For WBLS, its performance depends on the weighted penalty factor which needs to be specified in advance. In addition, the abandonment of the connections between the input layer and the feature layer may loss some interesting proprieties \cite{BLS2018, BLS_U, BLS_F, BLS_SM, BLS_R}, and even makes WBLS fall into some common pitfalls discussed in \cite{ELM_P}.
As for RM-BLS, the random perturbation matrix is of great importance to promote the robustness of the algorithm, but how to design such random perturbation matrix is lack of guidance at present.
Therefore, to develop a more general BLS which is expected to remain the advantages of the standard BLS as possible while having the ability to suppress the adverse effects of outliers still needs more efforts.
During the past few years, an effcient Information Theoretic Learning (ITL) \cite{ITL} criterion called the maximum correntropy criterion (MCC) has been successfully applied to adaptive filters \cite{KRMC, QKMC, MCKF}, randomized learning machines  \cite{RSCN-MCC, ESN-MCC, MCC_MELM1, MCC_MELM2, MCC-RELM}, principal component analysis (PCA) \cite{MCC-PCA}, auto-encoder \cite{MCC_AE1, MCC_AE2}, common spatial patterns (CSP) \cite{MCC_CSP}, and many others. These successful applications demonstrate that MCC performs very well with outliers. In addition, according to \emph{Property 3} provided in \cite{P-MCC}, correntropy has the potential to capture both the second-order and higher-order statistical characteristics of errors when the Gaussian kernel is used. With an appropriate setting of kernel size, the second-order statistical characteristics of errors can be dominant, which makes correntropy based optimization criterion also become a suitable choice for Gaussian noise or noise free environment. Inspired by the successful applications and attractive features of correntropy, we adopt it to train the output weights of BLS. Our main contributions are summarized as follows:
\begin{itemize}
 \item By using an MCC based fixed-point iteration algorithm to train the output weights of BLS, we propose a correntropy based BLS (C-BLS). The new method is robust to outliers, and has the potential to achieve comparable performance to the standard BLS in Gaussian noise or noise free environment.
 \item Three alternative incremental learning algorithms that are derived from a weighted regularized least-squares solution rather than pseudoinverse formula are provided.
     These algorithms ensure that the system can be remodeled quickly without the entire retraining process from the beginning when some new samples arrive or the network deems to be expanded.
 \item To test the effectiveness of the
proposed methods comprehensively, various regression and classification applications are provided for performance evaluation.
\end{itemize}

The remainder of the paper is organized as follows. In Section \ref{Sec2}, we give a brief review of BLS. In Section \ref{Sec3}, the correntropy is introduced, and based on correntropy, we propose the C-BLS and its incremental learning algorithms.
Section \ref{Sec4} presents experiment results on various regression and classification applications to demonstrate the performance of the proposed methods. At last, the conclusion is made in section \ref{Sec5}.

\section{Broad Learning System} \label{Sec2}
The basic idea of BLS comes from random vector functional-link neural networks (RVFLNN) \cite{FLNN1994, FLNN1999}, but the direct connections between the input layer and the output layer of RVFLNN are replaced by a set of general mapped features, and the system can be flatted in the wide sense by the enhancement nodes. Such deformation leads to some interesting properties and even makes BLS outperform several deep structure based learning methods \cite{BLS2018, BLS_U}.
\subsection{Basic Structure and Training Algorithm} \label{Sec2-1}
\begin{figure}[htbp]
  \centering
  \includegraphics[width=8.5cm]{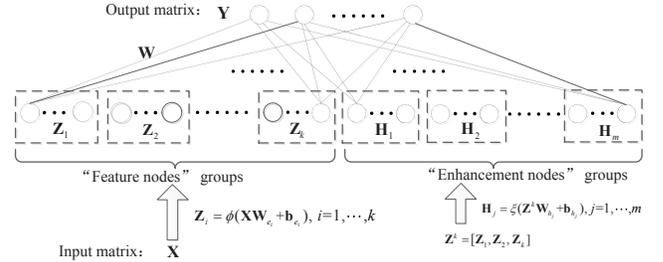}
  \caption{The basic architecture of broad learning system.}
  \label{fig1}
\end{figure}

Fig.~\ref{fig1} shows the basic architecture of BLS \cite{BLS2018}.
Herein, $\textbf{X}=[\textbf{x}_{1}^{T}, \textbf{x}_{2}^{T},\cdots,\textbf{x}_{N}^{T}]^{T}\in\mathbb{R}^{N\times M}$ and $\textbf{Y}=[\textbf{y}_{1}^{T}, \textbf{y}_{2}^{T},\cdots,\textbf{y}_{N}^{T}]^{T}\in\mathbb{R}^{N\times C}$ are respectively the input and the output matrices, where $N$ denotes the number of samples, $T$ represents the transpose operator, $M$ is the dimension of each input vector, and $C$ denotes the dimension of each output.

Based on $\textbf{X}$, $k$ groups of mapped features denoted as $\textbf{Z}_{1}, \textbf{Z}_{2}, \cdots, \textbf{Z}_{k}$ are firstly obtained by
\begin{equation}\label{E1}
{{\bf{Z}}_i} = {\phi _i}({\bf{X}}{{\bf{W}}_{e_i}} + {\boldsymbol{\beta}}_{e_i}) \in {\mathbb{R}^{N \times q}},\;\;i = 1,2, \cdots ,k
\end{equation}
where $\phi_{i}$ is usually a linear transformation; $q$ corresponds to the number of feature nodes in each group; $\textbf{W}_{e_i}\in {\mathbb{R}^{M \times q}}$ and $\boldsymbol{\beta}_{e_i}\in {\mathbb{R}^{N \times q}}$ are randomly generated weights and biases, respectively. In order to obtain the sparse representations of input data, they can be slightly fine-tuned by a sparse auto-encoder \cite{BLS2018}.
Concatenating all mapped features together, we have
\begin{equation}\label{E2}
 \textbf{Z}^{k}=[\textbf{Z}_{1}, \textbf{Z}_{2}, \cdots, \textbf{Z}_{k}]\in {\mathbb{R}^{N \times kq}}.
\end{equation}

Based on $\textbf{Z}^{k}$, $m$ groups of ``enhancement nodes" denoted as $\textbf{H}_{1}, \textbf{H}_{2}, \cdots, \textbf{H}_{m}$ are further obtained, that is
\begin{equation}\label{E3}
{{\bf{H}}_j} = {\xi _j}({{\bf{Z}}^k}{{\bf{W}}_{h_j}} + {\boldsymbol{\beta}}_{h_j}) \in {\mathbb{R}^{N \times r}},\;\;j = 1,2, \cdots ,m
\end{equation}
where $\xi_{j}$ is an activation function, such as $\xi_{j}(x)={\rm tanh}(x)$; $r$ corresponds to the number of enhancement nodes in each group; $\textbf{W}_{h_j}\in {\mathbb{R}^{kq \times r}}$ and $\boldsymbol{\beta}_{h_j}\in {\mathbb{R}^{N \times r}}$ are also randomly generated weights and biases, respectively. These ``enhancement nodes" can also be cascaded into one in the form of
\begin{equation}\label{E4}
 \textbf{H}^{m}=[\textbf{H}_{1}, \textbf{H}_{2}, \cdots, \textbf{H}_{m}]\in {\mathbb{R}^{N \times mr}}.
\end{equation}

By concatenating $\textbf{Z}^{k}$ and $\textbf{H}^{m}$ , we obtain
\begin{equation}\label{E5}
\textbf{U}=[\textbf{Z}^{k}, \textbf{H}^{m}]\in \mathbb{R}^{N\times L},
\end{equation}
where $L=kq + mr$. Clearly, $\textbf{U}$ is a new representation of the original input matrix $\textbf{X}$, and termed as the state matrix in \cite{BLS_R}.
Since all $\left\{ {{{\bf{W}}_{ei}},{{\boldsymbol{\beta}}_{ei}}} \right\}_{i = 1}^k$ and $\left\{ {{{\bf{W}}_{hj}},{{\boldsymbol{\beta}}_{hj}}} \right\}_{j = 1}^m$ are randomly
generated and remain unchanged after that, the learning task reduces to estimate the output weights $\textbf{W}$. This optimization problem can be modeled as to find the regularized least-squares solution of $\textbf{Y}=\textbf{U}\textbf{W}$, that is
\begin{align}\label{E7}
\mathop{\text{arg min}}\limits_{\textbf{W}}\left(\parallel\textbf{U}\textbf{W}-\textbf{Y}\parallel_{2}^{2}+\lambda \parallel\textbf{W}\parallel_{2}^{2} \right). 
\end{align}
Therefore, we have
\begin{align}\label{E8}
 \textbf{W}=(\textbf{U}^{T}\textbf{U}+\lambda \textbf{I})^{-1}\textbf{U}^{T}\textbf{Y}, 
\end{align}
in which $\textbf{I}$ denotes an identify matrix with proper dimensions, and $\lambda$ is a nonnegative constant for regularization. One should note that when $\lambda \rightarrow 0$, the solution in \eqref{E8} is equivalent to
\begin{align}\label{E9}
 \textbf{W}=\textbf{U}^{\dagger}\textbf{Y}, 
\end{align}
where $\textbf{U}^{\dagger}=\lim\limits_{\lambda \rightarrow 0}(\textbf{U}^{T}\textbf{U}+\lambda\textbf{I})^{-1}\textbf{U}^{T}$ denotes the pseudoinverse of $\textbf{U}$. Equation \eqref{E9} has been chosen as the main strategy in \cite{BLS2018} for finding the output weights $\textbf{W}$.
\subsection{Incremental Learning Algorithms for BLS}
We now give a brief introduction to the incremental learning algorithms of BLS. For simplicity, the subscripts of the feature mapping $\phi_{i}$ and the activation function $\xi_{j}$ will be omitted in the following, but one should note that $\phi_{i}$ can be selected
differently in practice as well as $\xi_{j}$. In addition, we denote $\textbf{X}(t)$ and $\textbf{Y}(t)$ as the current input matrix and the current output matrix, respectively. According to \eqref{E9}, the current output weights can be obtained by
\begin{align}\label{EBLSI1}
 \textbf{W}(t)=\textbf{U}(t)^{\dagger}\textbf{Y}(t). 
\end{align}
where $\textbf{U}(t)$ is the state matrix calculated according to \eqref{E1}-\eqref{E5}. Obviously, to derive the incremental learning algorithms of BLS, we need to determine the new forms of $\textbf{U}(t)$ and $\textbf{Y}(t)$.
\subsubsection{Increment of New Samples}
When $N_0$ new samples $\{\textbf{x}_{i},\textbf{y}_{i}\}_{i=N+1}^{N+N0}$ arrive, the increased input matrix and output matrix can be expressed by $\textbf{X}_{\alpha}=[\textbf{x}_{N+1}^{T},\cdots,\textbf{x}_{N+N_0}^{T}]^{T}\in\mathbb{R}^{N_0\times M}$ and $\textbf{Y}_{\alpha}=[\textbf{y}_{N+1}^{T}, \cdots,\textbf{y}_{N+N_0}^{T}]^{T}\in\mathbb{R}^{N_0\times C}$, respectively. The new state matrix and the output matrix are therefore obtained by
\begin{align}\label{EBLSI10}
\textbf{U}(t+1)=\left[
                  \begin{array}{c}
                    \textbf{U}(t) \\
                     \textbf{U}_{\alpha} \\
                  \end{array}
                \right],
~
\textbf{Y}(t+1)=\left[
                  \begin{array}{c}
                    \textbf{Y}(t) \\
                   \textbf{Y}_{\alpha} \\
                  \end{array}
                \right],
\end{align}
where $\textbf{U}_{\alpha}=[\textbf{Z}_{\alpha}^{k},\textbf{H}_{\alpha}^{m}]\in \mathbb{R}^{N_0 \times L}$, and
\begin{align}\label{EBLSI11}
\textbf{Z}_{\alpha}^{k}=\left[\phi(\textbf{X}_{\alpha}\textbf{W}_{e_{1}}+\boldsymbol{\beta}_{e_{1}}),\cdots,\phi(\textbf{X}_{\alpha}\textbf{W}_{e_{k}}+\boldsymbol{\beta}_{e_{k}})\right],
\end{align}
\begin{align}\label{EBLSI12}
\textbf{H}_{\alpha}^{m}=\left[\xi(\textbf{Z}_{\alpha}^{k}\textbf{W}_{h_{1}}+\boldsymbol{\beta}_{h_{1}}),\cdots,\xi(\textbf{Z}_{\alpha}^{k}\textbf{W}_{h_{m}}+\boldsymbol{\beta}_{h_{m}})\right],
\end{align}
According to \cite{BLS2018, FLNN1999}, the pseudoinverse
of $\textbf{U}(t+1)$ in \eqref{EBLSI10} can be calculated by
\begin{align}\label{ECI3}
{\textbf{U}(t+1)}^{\dagger} = \left[ {\textbf{U}(t)}^{\dagger}-\textbf{B}\textbf{D},~ \textbf{B}\right],
\end{align}
with
\begin{align}\label{EBLSI14}
\begin{array}{l}
\textbf{D}={\textbf{U}_{\alpha}}\textbf{U}(t)^{\dagger}\\
\textbf{B} = \left\{ \begin{array}{l}
\textbf{C}^{\dagger},\;\;\;\;\;\;\;\;\;\;\;\; \;\;\;\;\;\;\;\;\;\;\;\; \;\;\;\;\;\;\;\;\;\;\;\;\;\;{\rm if}\;\textbf{C} \neq \textbf{0}\\
(\textbf{I}+\textbf{D}\textbf{D}^{T})^{-1}\textbf{U}(t)^{\dagger}{\textbf{D}}^T,\;\;\;{\rm if}\;\textbf{C}= \textbf{0}
\end{array} \right. \\
\textbf{C}={\textbf{U}_{\alpha}}-{\textbf{D}}{\textbf{U}(t)} .
\end{array}
\end{align}
Correspondingly, the update equation for the output weights has the following form
\begin{align}\label{EBLSI13}
\textbf{W}(t+1)&=\textbf{U}(t\!+\!1)^{\dagger}\textbf{Y}(t\!+\!1)\nonumber\\
&=\textbf{W}(t)+\textbf{B}\left({\textbf{Y}_{\alpha}}-{\textbf{U}_{\alpha}}\textbf{W}(t)\right).
\end{align}

\subsubsection{Increment of Enhancement Mapping Nodes} \label{IE}
When $p$ new enhancement nodes are inserted, the state matrix changes to
\begin{align}\label{EBLSI14}
\textbf{U}(t\!+\!1)\!=\!\left[\textbf{U}(t), \xi(\textbf{Z}^{k}\textbf{W}_{h_{m\!+\!1}}\!+\!\boldsymbol{\beta}_{h_{m\!+\!1}})\right],
\end{align}
where $\textbf{W}_{h_{m+1}}\in {\mathbb{R}^{kq \times p}}$ and $\boldsymbol{\beta}_{h_{m+1}}\in {\mathbb{R}^{N \times p}}$ are randomly generated weights and biases, respectively.
The pseudoinverse of $\textbf{U}(t\!+\!1)$ in \eqref{EBLSI14} can be calculated in the following way \cite{BLS2018}
\begin{align}\label{EBLSI15}
{\textbf{U}(t+1)}^{\dagger} = \left[ {\begin{array}{*{20}{c}}
{\textbf{U}(t+1)}^{\dagger}-\textbf{D}\textbf{B}^{T}\\
\textbf{B}^{T}
\end{array}} \right],
\end{align}
with
\begin{align}\label{EBLSI16}
\begin{array}{l}
\textbf{D}=\textbf{U}(t)^{\dagger}\xi(\textbf{Z}^{k}\textbf{W}_{h_{m+1}}+\boldsymbol{\beta}_{h_{m+1}})\\
\textbf{B}^{T} = \left\{ \begin{array}{l}
\textbf{C}^{\dagger},\;\;\;\;\;\;\;\;\;\;\;\; \;\;\;\;\;\;\;\;\;\;\;\; \;\;\;\;\;\;\;\;{\rm if}\;\textbf{C} \neq \textbf{0}\\
(\textbf{I}+\textbf{D}^{T}\textbf{D})^{-1}\textbf{D}^{T}\textbf{U}(t)^{\dagger},\;\;\;{\rm if}\;\textbf{C}= \textbf{0}
\end{array} \right.\\
\textbf{C}=\xi(\textbf{Z}^{k}\textbf{W}_{h_{m+1}}+\boldsymbol{\beta}_{h_{m+1}})-\textbf{U}(t)\textbf{D}.
\end{array}
\end{align}
Since $\textbf{Y}(t+1)=\textbf{Y}(t)$, the output weights in this case are therefore updated by
\begin{align}\label{EBLSI17}
 \textbf{W}(t+1)\!=\!\textbf{U}(t\!+\!1)^{\dagger}\textbf{Y}(t\!+\!1)\!=\!\left[ {\begin{array}{*{20}{c}}
 \textbf{W}(t)-\textbf{D}\textbf{B}^{T}\textbf{Y}(t)\\
\textbf{B}^{T}\textbf{Y}(t)
\end{array}} \right]. 
\end{align}

\subsubsection{Increment of Feature Mapping Nodes}
When the $(k+1)$th group of feature nodes are inserted, we have
\begin{align}\label{EBLSI5}
\textbf{U}(t+1)=\left[\textbf{U}(t), \textbf{Z}_{k+1}, \textbf{H}_{\text{ex}_m} \right],
\end{align}
with
\begin{equation}\label{EBLSI6}
{{\bf{Z}}_{k+1}} = {\phi}({\bf{X}}{{\bf{W}}_{e_{k+1}}} + {{\boldsymbol{\beta}}_{e_{k+1}}}),
\end{equation}
\begin{equation}\label{EBLSI7}
\resizebox{0.88\hsize}{!}
{$
\textbf{H}_{\text{ex}_m}\!=\!\left[{\xi}({{\bf{Z}}_{k\!+\!1}}{{\bf{W}}_{\text{ex}_1}}\!+\!{{\boldsymbol{\beta}}_{\text{ex}_1}}),\!\cdots\!,{\xi}({{\bf{Z}}_{k\!+\!1}}{{\bf{W}}_{\text{ex}_m}}\!+\!{{\boldsymbol{\beta}}_{\text{ex}_m}})\right],
$}
\end{equation}
where $\{{\bf{W}}_{\text{ex}_i}, {\boldsymbol{\beta}}_{\text{ex}_i}\}_{i=1}^{m}$ are also randomly generated weights and
biases that connect new feature nodes to the
enhancement nodes. With a similar procedure used from \eqref{EBLSI15} to \eqref{EBLSI17}, the new output weights here can be calculated by
\begin{align}\label{EBLSI8}
 \textbf{W}(t+1)=\left[ {\begin{array}{*{20}{c}}
 \textbf{W}(t)-\textbf{D}\textbf{B}^{T}\textbf{Y}(t)\\
\textbf{B}^{T}\textbf{Y}(t)
\end{array}} \right], 
\end{align}
with
\begin{align}\label{EBLSI9}
\begin{array}{l}
\textbf{D}=\textbf{U}(t)^{\dagger}\left[\textbf{Z}_{k+1}, \textbf{H}_{\text{ex}_m} \right]\\
\textbf{B}^{T} = \left\{ \begin{array}{l}
\textbf{C}^{\dagger},\;\;\;\;\;\;\;\;\;\;\;\; \;\;\;\;\;\;\;\;\;\;\;\; \;\;\;\;\;\;\;\;{\rm if}\;\textbf{C} \neq \textbf{0}\\
(\textbf{I}+\textbf{D}^{T}\textbf{D})^{-1}\textbf{D}^{T}\textbf{U}(t)^{\dagger},\;\;\;{\rm if}\;\textbf{C}= \textbf{0}
\end{array} \right.\\
\textbf{C}=\left[\textbf{Z}_{k+1}, \textbf{H}_{\text{ex}_m} \right]-\textbf{U}(t)\textbf{D}.
\end{array}
\end{align}

\emph{Remark 1}: When some new samples arrive or some new nodes are involved, the above three incremental learning algorithms can update the output weights of BLS without needing to
run a complete training cycle, which ensures that the system can be remodeled quickly. However, these incremental learning algorithms require the regularization factor to tend to zero,
so that the regularized least squares solution can well approximate
the pseudo-inverse. This is, of course, not always the good
choice, since the regularization factor plays an important role
to improve the model's generalization ability in many practical
applications. In the next section, several more general incremental learning algorithms under the BLS architecture will be provided.

\section{Correntropy Based Broad Learning System} \label{Sec3}
Although BLS has so many attractive features, its dependence on the second order statistical
characteristics of errors makes it not a suitable choice in complicated noise environments, especially when data are disturbed by some outliers \cite{BLS_L1}. To offer a robust version of BLS, we introduce in this section the concept of correntropy, and based on correntropy, the C-BLS and its incremental learning algorithms are developed.

\subsection{Correntropy}
Correntropy \cite{P-MCC} is a local similarity measure between two arbitrary random variables $X$ and $Y$, defined by
\begin{align}\label{EC1}
  {V_\sigma }(X,Y) & =E\left[{{\kappa _\sigma }(X,Y)}\right] ,
\end{align}
where $E(\cdot)$ denotes the expectation operator, and ${\kappa _\sigma }( \cdot , \cdot )$ is a Mercer kernel \cite{MERCER} controlled by the kernel size $\sigma$. Without loss of generality, the Gaussian kernel defined as ${\kappa _\sigma }(x,y) = \frac{1}{\sqrt{2\pi}\sigma}\exp ( - \frac{{{{\left\| {x - y} \right\|}^2}}}{{2{\sigma ^2}}})
$ will be a default choice in this paper. By using the Taylor series expansion to \eqref{EC1}, we have
\begin{align}\label{EC2}
  {V_\sigma }(X,Y) & =\frac{1}{\sqrt{2\pi}\sigma}\sum\limits_{n=1}^{\infty}\frac{(-1)^{n}}{2^{n}n!}E\left[\frac{(X-Y)^{2n}}{\sigma^{2n}}\right].
\end{align}
Clearly, correntropy can be viewed as a weighted sum of all
even moments of $X-Y$, and the
weights of the second and higher-order moments are controlled by the kernel
size $\sigma$. As $\delta$ increases, the high-order moments decay faster. Hence,
the second-order moment has the chance to be dominant with a large $\sigma$.
In practice, the data distribution is usually unknown and only a finite number of samples $\{ ({{x}_i},{y_i})\} _{i = 1}^{{N}}$ are available, resulting in the sample estimator of correntropy to be
\begin{equation}\label{EC3}
  {\hat V_{\sigma }}(X,Y) = \frac{1}{N}\sum\limits_{i = 1}^N {{\kappa _\sigma }({x_i},{y_i})}.
\end{equation}
In signal processing and machine learning fields, it is usually to estimate an unknown parameter $\omega$ (such as the weight vector of an adaptive filter) by maximizing the
correntropy between the desired signal $Y$ and its estimated value $\hat{Y}$, i. e.,
\begin{align}\label{EC4}
\mathop{\text{arg max}}\limits_{\omega} {\hat V_{\sigma }}(Y,\hat{Y}) ,
\end{align}
This optimization criterion is called MCC. Unlike the well known MMSE criterion which is sensitive to outliers, MCC has been proven to be very robust for parameters estimation in complicated noise environments \cite{MCC_SINGLE, MCC_MULTI, MCC-ADMM, FR-MCC, KRMC, QKMC, MCCR}.

\subsection{Basic Training Algorithm for C-BLS}
Similar to the standard BLS, the state matrix $\textbf{U}$ in the proposed method can be constructed though a series of feature mappings and enhancement transformations that have been described in \eqref{E1}-\eqref{E5}. However, more powerful feature mapping strategies, such as convolution-pooling operation \cite{BLS_U}, neuro-fuzzy model \cite{BLS_F}, and structured manifold learning technology \cite{BLS_SM}, are also feasible. Thus, the optimization model that combines BLS and MCC can be formulated by
\begin{align}\label{EM1}
\mathop{\text{arg max}}\limits_{\textbf{W}} \left(\sum\limits_{i = 1}^N {\exp ( - \frac{{{{\left\| {{\textbf{u}_i} \textbf{W}  - {\textbf{y}_i}} \right\|}_{2}^2}}}{{2{\sigma ^2}}})}-\frac{\lambda}{2}\parallel\textbf{W}\parallel_{2}^{2}\right) ,
\end{align}
where $\textbf{u}_{i}\in \mathbb{R}^{L}$ denotes the $i$th row of $\textbf{U}$. For implicity, we denote $J(\textbf{W})=\sum_{i = 1}^N {\exp ( - \frac{{{{\left\| {{\textbf{u}_i} \textbf{W}  - {\textbf{y}_i}} \right\|}_{2}^2}}}{{2{\sigma ^2}}})}-\frac{\lambda}{2}\parallel\textbf{W}\parallel_{2}^{2}$, and then \eqref{EM1} can be rewritten as
\begin{align}\label{EM2}
\mathop{\text{arg max}}\limits_{\textbf{W}} J(\textbf{W}).
\end{align}
Taking the gradient of $J(\textbf{W})$ with respect to $\textbf{W}$, we have
\begin{align}\label{EM3}
\resizebox{0.88\hsize}{!}
{$
\frac{{\partial J}(\textbf{W})}{{\partial \textbf{W} }} \!=\! \!-\!\frac{1}{{{\sigma ^2}}}\sum\limits_{i \!= \!1}^N {\textbf{u}_{i}^T\exp ( \!-\! \frac{{{{\left\| {\textbf{u}_{i}\textbf{W}  \!-\! {\textbf{y}_i}} \right\|}_{2}^2}}}{{2{\sigma ^2}}})} (\textbf{u}_{i}\textbf{W}  \!-\! {\textbf{y}_i})\!-\!\lambda\textbf{W}
$}.
\end{align}
The matrix form of \eqref{EM3} can be expressed by
\begin{equation}\label{EM4}
  \frac{{\partial J}(\textbf{W})}{{\partial \textbf{W} }} =  - \frac{1}{{{\sigma ^2}}}{\textbf{U}^T}\boldsymbol{\Lambda}_{w}  (\textbf{U}\textbf{W}- \textbf{Y})-\lambda\textbf{W},
\end{equation}
with
\begin{equation}\label{EM5}
  \boldsymbol{\Lambda}_{w}  \!=\! \left[ {\begin{array}{*{20}{c}}
{\exp ( \!-\! \frac{{{{\left\| {{\textbf{u}_1}\textbf{W}\!-\!{\textbf{y}_1}} \right\|}_{2}^2}}}{{2{\sigma ^2}}})}&{}&{}\\
{}& \!\ddots\! &{}\\
{}&{}&{\exp ( - \frac{{{{\left\| {{\textbf{u}_N}\textbf{W}\!-\!{\textbf{y}_N}} \right\|}_{2}^2}}}{{2{\sigma ^2}}})}
\end{array}} \right].
\end{equation}
By setting $\frac{{\partial J}(\textbf{W})}{{\partial \textbf{W} }}$ to zero, the solution of $\textbf{W}$ can be
written in the following form
\begin{align}\label{EM6}
  \textbf{W}  &= {({\textbf{U}^T}\boldsymbol{\Lambda}_{w} \textbf{U}+\gamma\textbf{I})^{ - 1}}{\textbf{U}^T}\boldsymbol{\Lambda}_{w}  \textbf{Y},
\end{align}
where $\gamma=\lambda\sigma^{2}$. Obviously, $\boldsymbol{\Lambda}_{w} $ is the function of $\textbf{W}$. Hence, \eqref{EM6} is actually a fixed-point equation which can be described by
\begin{equation}\label{EM7}
  \textbf{W}  = f(\textbf{W} ),
\end{equation}
with
\begin{align}\label{EM77}
  f(\textbf{W} )  &= {({\textbf{U}^T}\boldsymbol{\Lambda}_{w}  \textbf{U}+\gamma\textbf{I})^{ - 1}}{\textbf{U}^T}\boldsymbol{\Lambda}_{w}  \textbf{Y}.
\end{align}
Referring to the widely used fixed-point iteration method \cite{FP-Theory, FP-MCCT, FP-MCC-MEE, FP-KMP}, we can solve $\textbf{W}$ by the following iteration way
\begin{equation}\label{EM8}
  {{ {\textbf{W}} }{(t + 1)}} = f({{ {\textbf{W}} }{(t)}}),
\end{equation}
where ${{ {\textbf{W}} }{(t)}}$ denotes the solution at iteration $t$. Let $\varepsilon$ denote termination
tolerance, the stopping criterion can be set as $\|\textbf{W}(t\!+\!1)\big)\!-\!\textbf{W}(t)\|_{2}^{2}\!<\!\varepsilon$.
%
According to the work done in \cite{FP-MCCT}, the convergence of the fixed-point iteration method under the MCC can be guaranteed if the kernel size $\sigma$ is appropriately chosen. In the following experiments, the grid search method will be adopted to determine $\sigma$ and other parameters of C-BLS, so as to ensure its convergence and also make it approach its optimal performance as possible.

Finally, the proposed C-BLS is summarized in \textbf{Algorithm \ref{Alg1}}.
\begin{algorithm}[htp]\label{Alg1}
\caption{Correntropy Based Broad Learning System}
\begin{algorithmic}
\STATE{\rule{-0.5cm}{0.4cm}}\textbf{Input}: Training set $\{\textbf{X},\textbf{Y}\}$.
\STATE{\rule{-0.5cm}{0.4cm}}\textbf{Output}: Output weight matrix $\textbf{W}$.
\STATE{\rule{-0.3cm}{0.4cm}}1. \textbf{Parameters setting}: network parameters $k$, $q$, $m$, $r$, \\
 \  regularization parameter $\gamma$, kernel size $\sigma$, termination \\
   \ tolerance $\varepsilon$, maximum iteration number $T$. 
\STATE{\rule{-0.3cm}{0.4cm}}2. \textbf{Initialization}: set $\textbf{W}_{0}$ and construct the state matrix \\
\ $\textbf{U}$ according to \eqref{E1}-\eqref{E5}.
\STATE{\rule{-0.3cm}{0.4cm}}3. \textbf{for} $t = 1, ...,T$ \textbf{do}
\STATE{\rule{-0.3cm}{0.4cm}}4. \ \ \  Compute $\boldsymbol{\Lambda}_{w}$ according to \eqref{EM5};
\STATE{\rule{-0.3cm}{0.4cm}}6. \ \ \  Update $\textbf{W}$ according to \eqref{EM6};
\STATE{\rule{-0.3cm}{0.4cm}}7. \ \ \  Until $\|\textbf{W}(t\!+\!1)\big)\!-\!\textbf{W}(t)\|_{2}^{2}\!<\!\varepsilon$.
\STATE{\rule{-0.3cm}{0.4cm}}8. \textbf{end for}
\end{algorithmic}
\end{algorithm}

\emph{Remark 2}: Compared with \eqref{E8} for the original BLS, \eqref{EM6} has an additive weighted diagonal matrix $\boldsymbol{\Lambda}_{w} $, whose the $i$th diagonal element is controlled by the kernel size $\sigma$ as well as the difference between $\textbf{y}_i$ and its estimation $\hat{\textbf{y}}_{i}=\textbf{u}_{i}\textbf{W}$. It can be verified that when $\sigma \rightarrow \infty$, \eqref{EM6} will reduce to the solution of the standard BLS.
This makes C-BLS can, at least, achieve the comparable
performance to the standard BLS. In addition, by appropriately setting
the values of $\sigma$, C-BLS has the potential to weaken
the negative effects of outliers. For example, when the $i$th sample is polluted by outlier, there will be in general a large difference  between $\textbf{y}_{i}$ and $\hat{\textbf{y}}_{i}$, denoted as $\nu=\parallel\textbf{y}_{i}-\hat{\textbf{y}}_{i}\parallel_2^{2}$. With appropriately setting of $\sigma$, such as $\sigma \rightarrow 0$, we have $\text{exp}(- \frac{\nu}{2\sigma^2}) \rightarrow 0$, which makes the outlier not have a big impact on the training process.

\emph{Remark 3}: The update equations of the proposed C-BLS is somewhat similar to the W-BLS proposed in \cite{WBLS}.
However, there are several dissimilarities between them, including: 1) C-BLS is proposed from the Information Theoretic Learning （ITL) \cite{ITL} perspective while W-BLS is proposed from the application of industrial process;  2) the weighted operator in C-BLS is a successive result derived from MCC while the weighted operator in W-BLS is an additional hyperparameter needed to be specified in advance; 3) C-BLS remains the connections
between the input layer and the feature layer and hence can be easily combined with some existing feature mapping technologies \cite{BLS2018, BLS_U, BLS_F, BLS_SM} while W-BLS abandons such connections. 

\subsection{Incremental Learning Algorithms for C-BLS}

To derive the incremental learning algorithms of C-BLS, we also use ${\bf{X}}(t)$ and ${\bf{Y}}(t)$ to denote the current input matrix and the current output matrix, respectively. According to \eqref{EM6}, we therefore get that
\begin{align}\label{EI1}
  \textbf{W}(t) &= {[{\textbf{U}(t)^T}\boldsymbol{\Lambda}_{w(t)} \textbf{U}(t)+\gamma\textbf{I}]^{ - 1}}{\textbf{U}(t)^T}\boldsymbol{\Lambda}_{w(t)} \textbf{Y}(t)\nonumber\\
  &= {({\textbf{U}_{w(t)}}^T \textbf{U}_{w(t)}+\gamma\textbf{I})^{ - 1}}{\textbf{U}_{w(t)}}^T \textbf{Y}_{w(t)},
\end{align}
where ${\textbf{U}_{w(t)}}=\sqrt{\boldsymbol{\Lambda}_{w(t)}}\textbf{U}(t)$ is the weighted state matrix,   $\textbf{Y}_{w(t)}=\sqrt{\boldsymbol{\Lambda}_{{w}(t)}}\textbf{Y}(t)$ corresponds to the weighted output matrix, and $\boldsymbol{\Lambda}_{w(t)}$ is calculated by
\begin{equation}\label{EI2}
\resizebox{0.88\hsize}{!}
{$
  \boldsymbol{\Lambda}_{w(t)}\!=\!\left[ {\begin{array}{*{20}{c}}
{\exp ( \!-\! \frac{{{{\left\| {{\textbf{u}_1}\textbf{W}(t)\!-\!{\textbf{y}_1}} \right\|}_{2}^2}}}{{2{\sigma ^2}}})}&{}&{}\\
{}& \!\ddots\! &{}\\
{}&{}&{\exp(\!-\!\frac{{{{\left\| {{\textbf{u}_N}\textbf{W}(t)\!-\!{\textbf{y}_N}} \right\|}_{2}^2}}}{{2{\sigma ^2}}})}
\end{array}} \right].
$}
\end{equation}
For ease of representation, we define $\textbf{R}_{w(t)}={\textbf{U}_{w(t)}^T} \textbf{U}_{w(t)}+\gamma\textbf{I}$ and  $\textbf{P}_{w(t)}={\textbf{U}_{w(t)}^T} \textbf{Y}_{w(t)}$. Hence, \eqref{EI1} can be written as
\begin{align}\label{EI4}
  \textbf{W}(t) &= \textbf{R}_{w(t)}^{-1}\textbf{P}_{w(t)}.
\end{align}
\subsubsection{Increment of new samples}
Assume that $N_0$ new samples $\{\textbf{x}_{i},\textbf{y}_{i}\}_{i=N+1}^{N+N0}$ are available. We first denote $\textbf{X}_{\alpha}=[\textbf{x}_{N+1}^{T},\cdots,\textbf{x}_{N+N_0}^{T}]^{T}\in\mathbb{R}^{N_0\times M}$ and $\textbf{Y}_{\alpha}=[\textbf{y}_{N+1}^{T}, \cdots,\textbf{y}_{N+N_0}^{T}]^{T}\in\mathbb{R}^{N_0\times C}$. 
Then, the weighted state matrix and output matrix can be obtained by
\begin{align}\label{EI5}
\textbf{U}_{w(t+1)}\approx\left[
                  \begin{array}{c}
                    \textbf{U}_{w(t)} \\
                   \textbf{U}_{w(t)}^{\alpha} \\
                  \end{array}
                \right],
~
\textbf{Y}_{w(t+1)}\approx\left[
                  \begin{array}{c}
                    \textbf{Y}_{w(t)} \\
                   \textbf{Y}_{w(t)}^{\alpha} \\
                  \end{array}
                \right],
\end{align}
where $\textbf{U}_{w(t)}^{\alpha}=[\textbf{Z}_{w(t)}^{k},\textbf{H}_{w(t)}^{m}]\in \mathbb{R}^{N_0 \times L}$ and $\textbf{Y}_{w(t)}^{\alpha}=\sqrt{\boldsymbol{\Lambda}_{w(t)}^{\alpha}}\textbf{Y}_{\alpha}\in \mathbb{R}^{N_0 \times C}$ with
\begin{equation}\label{EI6}
\resizebox{0.89\hsize}{!}
{$
  \boldsymbol{\Lambda}_{{w}(t)}^{\alpha}\!=\!\left[ {\begin{array}{*{20}{c}}
{\exp ( \!-\! \frac{{{{\left\| {{\textbf{u}_{N\!+\!1}}\textbf{W}(t)\!-\!{\textbf{y}_{N\!+\!1}}} \right\|}_{2}^2}}}{{2{\sigma ^2}}})}&{}&{}\\
{}& \!\ddots\! &{}\\
{}&{}&{\exp(\!-\!\frac{{{{\left\| {{\textbf{u}_{N\!+\!N_0}}\textbf{W}(t)\!-\!{\textbf{y}_{N\!+\!N_0}}} \right\|}_{2}^2}}}{{2{\sigma ^2}}})}
\end{array}} \right],
$}
\end{equation}
\begin{align}\label{EI7}
\resizebox{0.88\hsize}{!}
{$
\textbf{Z}_{w(t)}^{k}\!=\!\sqrt{\boldsymbol{\Lambda}_{w(t)}^{\alpha}}\left[\phi(\textbf{X}_{\alpha}\textbf{W}_{e_{1}}\!+\!\boldsymbol{\beta}_{e_{1}}),\!\cdots\!,\phi(\textbf{X}_{\alpha}\textbf{W}_{e_{k}}\!+\!\boldsymbol{\beta}_{e_{k}})\right],
$}
\end{align}
\begin{align}\label{EI8}
\resizebox{0.88\hsize}{!}
{$
\textbf{H}_{w(t)}^{m}=\sqrt{\boldsymbol{\Lambda}_{w(t)}^{\alpha}}\left[\xi(\textbf{Z}_{\alpha}^{k}\textbf{W}_{h_{1}}+\boldsymbol{\beta}_{h_{1}}),\cdots,\xi(\textbf{Z}_{\alpha}^{k}\textbf{W}_{h_{m}}+\boldsymbol{\beta}_{h_{m}})\right].
$}
\end{align}
Substituting $\textbf{U}_{w(t+1)}$ and $\textbf{Y}_{w(t+1)}$ into \eqref{EI4}, we have
\begin{align}\label{EI9}
  \textbf{W}(t+1) &= \textbf{R}_{w(t+1)}^{-1}\textbf{P}_{w(t+1)},
\end{align}
where
\begin{align}\label{EI10}
 \textbf{R}_{w(t+1)} &=\textbf{U}_{w(t+1)}^T\textbf{U}_{w(t+1)}+\gamma\textbf{I} \nonumber\\
 &\approx{\left[ {\begin{array}{*{20}{c}}
\textbf{U}_{w(t)}\\
\textbf{U}_{w(t)}^{\alpha}
\end{array}}\right]^T}
\left[ {\begin{array}{*{10}{c}}
\textbf{U}_{w(t)}\nonumber\\
\textbf{U}_{w(t)}^{\alpha}
\end{array}} \right]+\gamma\textbf{I}\nonumber\\
&=\textbf{R}_{w(t)} + {\textbf{U}_{w(t)}^{\alpha}}^T{\textbf{U}_{w(t)}^{\alpha}},
\end{align}
and
\begin{align}\label{EI11}
\textbf{P}_{w(t+1)} &= \textbf{U}_{w(t+1)}^T \textbf{Y}_{w(t+1)}\nonumber\\
&\approx{\left[ {\begin{array}{*{20}{c}}
\textbf{U}_{w(t)}\nonumber\\
\textbf{U}_{w(t)}^{\alpha}
\end{array}} \right]^T}
\left[ {\begin{array}{*{20}{c}}
\textbf{Y}_{w(t)}\nonumber\\
\textbf{Y}_{w(t)}^{\alpha}
\end{array}} \right]\nonumber\\
&=\textbf{P}_{w(t)} + {\textbf{U}_{w(t)}^{\alpha}}^T\textbf{Y}_{w(t)}^{\alpha}.
\end{align}

By using the matrix inverse lemma
\begin{equation}\label{EI12}
  (\textbf{A}+\textbf{BCD})^{-1}=\textbf{A}^{-1} - \textbf{A}^{-1}\textbf{B}(\textbf{C}^{-1}+\textbf{DA}^{-1}\textbf{B})^{-1}{\textbf{DA}}^{-1}
\end{equation}
with the definitions of $\textbf{A}=\textbf{R}_{w(t)}$, $\textbf{B}={\textbf{U}_{w(t)}^{\alpha}}^T$, $\textbf{C}=\textbf{I}$ and $\textbf{D}={\textbf{U}_{w(t)}^{\alpha}}$, we get
\begin{equation}\label{EI14}
\textbf{R}_{w(t+1)}^{-1} =
\textbf{R}_{w(t)}^{-1} - \textbf{R}_{w(t)}^{-1}{\textbf{U}_{w(t)}^{\alpha}}^T \textbf{S}_{w(t)}^{\alpha} {\textbf{U}_{w(t)}^{\alpha}}\textbf{R}_{w(t)}^{-1},
\end{equation}
where $\textbf{S}_{w(t)}^{\alpha}=[\textbf{I}+\textbf{U}_{w(t)}^{\alpha}\textbf{R}_{w(t)}^{-1}{\textbf{U}_{w(t)}^{\alpha}}^T]^{-1}$.
Substituting \eqref{EI11} and \eqref{EI14} into \eqref{EI9}, yields
\begin{equation}\label{EI15}
\resizebox{0.89\hsize}{!}
{$
\begin{array}{l}
\textbf{W}(t\!+\!1)  \!=\! \textbf{R}_{w(t+1)}^{-1}\textbf{P}_{w(t+1)}\\
 \ \ \ \ \ \ \ \ \ \ \ \!=\!\textbf{W}(t) \!+\! \textbf{R}_{w(t)}^{-1}{{\textbf{U}_{w(t)}^{\alpha}}}^T \textbf{S}_{w(t)}^{\alpha}\big({\textbf{Y}_{w(t)}^{\alpha}}\!-\!{\textbf{U}_{w(t)}^{\alpha}}\textbf{W}(t)\big).
\end{array}
$}
\end{equation}

Let $\textbf{C}_{w(t)}=\textbf{R}_{w(t)}^{-1}$. We obtain the final equations for updating $\textbf{W}(t+1)$ as follows
\begin{align}\label{EI16}
 & \textbf{S}_{w(t)}^{\alpha}\!=\![\textbf{I}+\textbf{U}_{w(t)}^{\alpha}\textbf{C}_{w(t)}{\textbf{U}_{w(t)}^{\alpha}}^T]^{-1}\nonumber\\
&\textbf{W}(t\!+\!1)\!=\!\textbf{W}(t) \!+\! {\textbf{C}_{w(t)}{\textbf{U}_{w(t)}^{\alpha}}}^T \textbf{S}_{w(t)}^{\alpha}\big({\textbf{Y}_{w(t)}^{\alpha}}\!-\!{\textbf{U}_{w(t)}^{\alpha}}\textbf{W}(t)\big)\nonumber\\
& \textbf{C}_{w(t\!+\!1)} \!=\!\textbf{C}_{w(t)} - \textbf{C}_{w(t)}{\textbf{U}_{w(t)}^{\alpha}}^T \textbf{S}_{w(t)}^{\alpha} {\textbf{U}_{w(t)}^{\alpha}}\textbf{C}_{w(t)}.
\end{align}
The main computational effort in \eqref{EI16} focuses on the calculation of $\textbf{S}_{w(t)}^{\alpha}$. Since only the latest $N_0$ samples as well as the previous $\textbf{C}_{w(t)}$ are involved to compute $\textbf{S}_{w(t)}^{\alpha}$, the corresponding computational cost is in general not burdensome.

\subsubsection{Increment of Enhancement Nodes}
Assume that $p$ enhancement nodes are inserted.
The weighted state matrix can be expressed by
\begin{align}\label{EI7}
\textbf{U}_{w(t\!+\!1)}\!\approx\!\left[\textbf{U}_{w(t)}, {\xi}_{w(t)}\right],
\end{align}
where ${\xi}_{w(t)}=\sqrt{\boldsymbol{\Lambda}_{w(t)}}\xi(\textbf{Z}^{k}\textbf{W}_{h_{m\!+\!1}}\!+\!\boldsymbol{\beta}_{h_{m\!+\!1}})
\in {\mathbb{R}^{N \times p}}$; $\textbf{W}_{h_{m+1}}\in {\mathbb{R}^{kq \times p}}$ and $\boldsymbol{\beta}_{h_{m+1}}\in {\mathbb{R}^{N \times p}}$ are randomly generated weights and biases, respectively.
With the approximation of $\textbf{Y}_{w(t+1)}\approx\textbf{Y}_{w(t)}$, the output weights in this case are obtained by
\begin{align}\label{EI18}
  \textbf{W}(t+1) &= \textbf{R}_{w(t+1)}^{-1}\textbf{P}_{w(t+1)},
\end{align}
with
\begin{align}\label{EI19}
\textbf{R}_{w(t+1)} &=\textbf{U}_{w(t+1)}^T\textbf{U}_{w(t+1)}+\gamma\textbf{I} \nonumber\\
 &\approx{\left[ {\begin{array}{*{20}{c}}
{\textbf{U}_{w(t)}^T}\\
{{\xi}_{w(t)}^T}
\end{array}}\right]}
[\textbf{U}_{w(t)},{\xi}_{w(t)}] +\gamma\textbf{I}\nonumber\\
&=\left[ {\begin{array}{*{20}{c}}
\textbf{R}_{w(t)}& \textbf{U}_{w(t)}^T  {\xi}_{w(t)} \\
{\xi}_{w(t)}^T \textbf{U}_{w(t)} &\gamma\textbf{I}+ {\xi}_{w(t)}^T {\xi}_{w(t)}
\end{array}} \right],
\end{align}
and
\begin{align}\label{EI20}
 {\textbf{P}_{w(t+1)}} &=\textbf{U}_{w(t+1)}^T{\textbf{Y}_{w(t+1)}} \nonumber\\
 &\approx{\left[ {\begin{array}{*{20}{c}}
{{\textbf{U}_{w(t)}^T}}\\
{{\xi}_{w(t)}^T}
\end{array}}\right]}
\textbf{Y}_{w(t)}\nonumber\\
&=\left[ {\begin{array}{*{20}{c}}
\textbf{P}_{w(t)} \\
{\xi}_{w(t)}^T \textbf{Y}_{w(t)}
\end{array}} \right],
\end{align}
The inverse matrix of $\textbf{R}_{w(t+1)}$ in \eqref{EI18} can be calculated by using the block matrix inversion lemma \cite{KAF}, which has the form of
\begin{equation}\label{EI20}
\resizebox{0.88\hsize}{!}{$
{\left[{\begin{array}{*{20}{c}}
{\bf{A}}&{\bf{B}}\\
{\bf{C}}&{\bf{D}}
\end{array}} \right]^{ - 1}} \!=\! \\
\left[ {\begin{array}{*{20}{c}}
{{{({\bf{A}} \!-\! {\bf{B}}{{\bf{D}}^{\!-\! 1}}{\bf{C}})}^{ \!-\! 1}}}&{ \!-\! {{\bf{A}}^{ \!-\! 1}}{\bf{B}}{{({\bf{D}} \!-\! {\bf{C}}{{\bf{A}}^{ \!-\! 1}}{\bf{B}})}^{ \!-\! 1}}}\\
\!-\! {({\bf{D}} - {\bf{C}}{{\bf{A}}^{ - 1}}{\bf{B}})^{ - 1}}{\bf{C}}{{\bf{A}}^{ - 1}}&{{{({\bf{D}} \!-\! {\bf{C}}{{\bf{A}}^{ \!-\! 1}}{\bf{B}})}^{ \!-\! 1}}}
\end{array}} \right]
$},
\end{equation}
where $\textbf{A}$ and $\textbf{D}$ are arbitrary reversible matrix blocks.
Let $\textbf{A}=\textbf{R}_{w(t)}$, $\textbf{B}= \textbf{U}_{w(t)}^T  {\xi}_{w(t)}$, $\textbf{C}= {\xi}_{w(t)}^T \textbf{U}_{w(t)}$, and $\textbf{D}=\gamma\textbf{I}+ {\xi}_{w(t)}^T {\xi}_{w(t)}$,
then we get
\begin{equation}\label{EI21}
\resizebox{0.88\hsize}{!}
{$
{\bf{R}}_{w(t + 1)}^{ - 1} = \left[ {\begin{array}{*{20}{c}}
{{\bf{R}}_{_{w(t)}}^{ - 1} + {{\bf{Z}}_{w(t)}}{{\bf{Q}}_{w(t)}}{\bf{Z}}_{w(t)}^T}&{ - {{\bf{Z}}_{w(t)}}{{\bf{Q}}_{w(t)}}}\\
{ - {{\bf{Q}}_{w(t)}}{{\bf{Z}}_{w(t)}^{T}}}&{{{\bf{Q}}_{w(t)}}}
\end{array}} \right]
$},
\end{equation}
where
\begin{equation} \label{EI211}
\begin{array}{l}
{{\bf{Z}}_{w(t)}} \!=\! {\bf{R}}_{_{w(t)}}^{ - 1}{\bf{U}}_{w(t)}^T{\xi _{w(t)}},\;\\
{\bf{Q}}_{w(t)} \!=\! {\left( {\gamma {\bf{I}} \!+\! \xi _{w(t)}^T{\xi _{w(t)}} \!-\! \xi _{w(t)}^T{{\bf{U}}_{w(t)}}\;{{\bf{Z}}_{w(t)}}} \right)^{ - 1}}.
\end{array}
\end{equation}
Substituting \eqref{EI20} and \eqref{EI21} into \eqref{EI18}, we have
\begin{equation}\label{EI22}
\resizebox{0.89\hsize}{!}
{$
\begin{array}{l}
{\bf{W}}(t{\rm{ + }}1) \!=\! {\bf{R}}_{w(t + 1)}^{ - 1}{{\bf{P}}_{w(t + 1)}}\\
\;\;\;\;\;\;\;\;\;\;\;\;\;\; \!=\! \left[ {\begin{array}{*{20}{c}}
{{\bf{W}}(t) - {{\bf{Z}}_{w(t)}}{{\bf{Q}}_{w(t)}}\xi _{w(t)}^T{\left( {{{\bf{Y}}_{w(t)}} - {{\bf{U}}_{w(t)}}{\bf{W}}(t)} \right)}}\\
{{{\bf{Q}}_{w(t)}}\xi _{w(t)}^T {\left( {{{\bf{Y}}_{w(t)}} - {{\bf{U}}_{w(t)}}{\bf{W}}(t)} \right)}}
\end{array}} \right].
\end{array}
$}
\end{equation}
\subsubsection{Increment of Feature Mapping Nodes} Assume that the $(k+1)$th group of feature nodes are inserted. The new weighted state matrix can be constructed by
\begin{align}\label{EI23}
\textbf{U}_{w(t+1)}\approx\left[\textbf{U}_{w(t)}, \vartheta_{w(t)} \right],
\end{align}
where $\vartheta_{w(t)}=[\textbf{Z}_{w(t)}, \textbf{H}_{w(t)}]$ with
\begin{equation}\label{EI24}
\textbf{Z}_{w(t)} = \sqrt{\boldsymbol{\Lambda}_{w(t)}}{\phi}({\bf{X}}{{\bf{W}}_{e_{k+1}}} + {{\boldsymbol{\beta}}_{e_{k+1}}}),
\end{equation}
\begin{equation}\label{EI25}
\resizebox{0.88\hsize}{!}
{$
\textbf{H}_{w(t)}\!=\!\sqrt{\boldsymbol{\Lambda}_{w(t)}}\left[{\xi}({{\bf{Z}}_{k\!+\!1}}{{\bf{W}}_{\text{ex}_1}}\!+\!{{\boldsymbol{\beta}}_{\text{ex}_1}}),\!\cdots\!,{\xi}({{\bf{Z}}_{k\!+\!1}}{{\bf{W}}_{\text{ex}_m}}\!+\!{{\boldsymbol{\beta}}_{\text{ex}_m}})\right].
$}
\end{equation}
Herein, $\{{\bf{W}}_{\text{ex}_i}, {\boldsymbol{\beta}}_{\text{ex}_i}\}_{i=1}^{m}$ are weights and
biases, which are also randomly generated. Similar to the case of increasing enhancement nodes, we get the update equation for the output weights as
\begin{equation}\label{EI25}
\resizebox{0.89\hsize}{!}
{$
\begin{array}{l}
{\bf{W}}(t{\rm{ + }}1) = {\bf{R}}_{w(t + 1)}^{ - 1}{{\bf{P}}_{w(t + 1)}}\\
\;\;\;\;\;\;\;\;\;\;\;\; =  \left[ {\begin{array}{*{20}{c}}
{{\bf{W}}(t) - {{\bf{Z}}_{w(t)}}{{\bf{Q}}_{w(t)}}\vartheta_{w(t)}^T{\left( {{{\bf{Y}}_{w(t)}} - {{\bf{U}}_{w(t)}}{\bf{W}}(t)} \right)}}\\
{{{\bf{Q}}_{w(t)}}\vartheta_{w(t)}^T {\left( {{{\bf{Y}}_{w(t)}} - {{\bf{U}}_{w(t)}}{\bf{W}}(t)} \right)}}
\end{array}} \right] ,
\end{array}
$}
\end{equation}
where
\begin{equation} \label{EI26}
\resizebox{0.89\hsize}{!}
{$
\begin{array}{l}
{{\bf{Z}}_{w(t)}} \!=\! {\bf{R}}_{_{w(t)}}^{ \!-\! 1}{\bf{U}}_{w(t)}^T{\vartheta _{w(t)}},\;\\
{\bf{Q}}_{w(t)} \!=\! {\left( {\gamma {\bf{I}} \!+\! \vartheta _{w(t)}^T{\vartheta _{w(t)}} \!-\! \vartheta _{w(t)}^T{{\bf{U}}_{w(t)}} {{\bf{Z}}_{w(t)}}} \right)^{ \!-\! 1}}\\
{\bf{R}}_{w(t + 1)}^{ - 1} = \left[ {\begin{array}{*{20}{c}}
{{\bf{R}}_{_{w(t)}}^{ - 1} + {{\bf{Z}}_{w(t)}}{{\bf{Q}}_{w(t)}}{\bf{Z}}_{w(t)}^T}&{ - {{\bf{Z}}_{w(t)}}{{\bf{Q}}_{w(t)}}}\\
{ - {{\bf{Q}}_{w(t)}}{{\bf{Z}}_{w(t)}^{T}}}&{{{\bf{Q}}_{w(t)}}}
\end{array}} \right] .
\end{array}
$}
\end{equation}

\emph{Remark 4}: It is clear that all the incremental learning algorithms
proposed in this section support the fast re-modeling,
since the main computational efforts in them focus on the
calculations of the additional part rather than the whole
body of weighted auto-correlation matrix $\textbf{R}_{w(t+1)}$. In addition, we don't restrict
that the regularization factor tends to zero in all derivation
processes, which makes them have the potential to get a better generalization in practice.
Thus, they can be viewed as general incremental learning
algorithms under the BLS architecture.

\section{Experiment Results} \label{Sec4}
In this section, the experiments are presented to demonstrate the desirable performance of the proposed C-BLS and its incremental learning algorithms. Except mentioned otherwise, all
the experiment results are obtained using MATLAB (R2017b) on
a 3.60-GHz machine with 16-GB RAM.
\subsection{Performance Evaluation of C-BLS}\label{Sec4A}
Thirteen benchmark datasets, seven of which are for regression and the other six are for classification, from the UCI repository \cite{UCI} are adopted here to valuate the performance of C-BLS. The details of these datasets are shown in Table~\ref{Tab1}, in which the symbol '$\#$' represents the value of a variable or the number of samples in a set, and the symbol '-' means that the dataset is used for regression only. To eliminate the influence of data scales, both input and output attributes for
regression datasets are normalized into the range $[0, 1]$, while for
classification datasets, only the input attributes are normalized into
the range $[-1, 1]$.
\begin{table}[htbp]
 \renewcommand\arraystretch{1.2}
\newcommand{\tabincell}[2]{\begin{tabular}{@{}#1@{}}#2\end{tabular}}
 \caption{Specification of the fourteen benchmark data sets. }
 \begin{center}
 \begin{tabular}{lcccccccc}
  \toprule
Dataset & $\#$Class & $\#$Attributes  & $\#$Training  & $\#$Testing\\
  \midrule
    Airfoil Self-Noise & - & 5 & 1002 & 501 \\
    Bodyfat & - & 14 & 168 & 84  \\
    Cleveland & - & 13 & 202 & 101  \\
    Forestfires & - & 12 & 345 & 172  \\
    Mortgage & - & 15 & 699 & 350 \\
    Quake & - & 3 & 1452 & 726 \\
    Weather Izmir & - & 9 & 974 & 487 \\
Balance Scale & 3 & 4 & 417 & 208\\
Chess & 2 & 36 & 2131 & 1065 \\
Diabetes & 2 & 8 & 512 & 256 \\
Ecoli & 8 & 7 & 222 & 114 \\
Horse Colic & 2 & 27 & 300 & 68 \\
Wireless & 4 & 7 & 1332 & 668 \\
 \bottomrule
 \end{tabular}
\label{Tab1}
 \end{center}
\end{table}

\subsubsection{Comparison with the Standard BLS}
First, we compare the performance of C-BLS with the standard BLS on the aforementioned thirteen datasets.
There are several common parameters for BLS and C-BLS, including the number of feature
nodes $N_f$, the number of mapping groups $N_m$, the number of enhancement nodes $N_e$, and the regularization factor $\gamma$.
Similar to \cite{BLS_U}, we simply set the regularization factor to $\gamma=2^{-30}$, and perform a grid search on the best combination of $N_f$, $N_w$ and $N_e$. In detail, the search ranges for $N_f$ is $[1,20]$ with the step-size of $2$, for $N_w$ is $[1,20]$ with the step-size of $1$, and for $N_e$ is $[1,200]$ with the step-size of $5$. As for the kernel size introduced in C-BLS, it is searched in the range of $\{2^{-5}, 2^{-4}, \cdots, 2^{4}, 2^{5}\}$.

Table~\ref{Tab2} and Table~\ref{Tab3} show the comparative results of BLS and C-BLS on regression and classification datasets, respectively. To reduce the random error caused by experiment itself, all the results have been averaged over $20$ Monte Carlo runs. It can be seen from Table~\ref{Tab2} that, except for the dataset ``Forestfires", C-BLS can always reach a smaller or the same testing root mean square error (RMSE) \cite{BLS_SM} compared with the standard BLS. Furthermore, the comparative results on Table~\ref{Tab2} shows that, even applied to classification tasks, C-BLS also have the ability to get the same or a higher testing accuracy on most of datasets. However, C-BLS in general needs more training time compared with the standard BLS, since there is no closed-solution under MCC.
To reduce the time consuming, one can adopt some existing technologies, like single-value-decomposition (SVD) technology \cite{BLS_L1, SVD1990} or just providing a good initialization \cite{MCKF} for speeding up the calculation. Due to space limitation, we don't give a through discussion here.

\begin{table*}[htbp]
 \renewcommand\arraystretch{1.2}
\newcommand{\tabincell}[2]{\begin{tabular}{@{}#1@{}}#2\end{tabular}}
 \caption{Performance comparison of BLS and C-BLS on regression benchmark data sets}
 \begin{center}
 \begin{tabular}{llccccccc}
  \toprule
Dataset & Algorithm  & Parameters & Training Time (s)   & Training RMSE & Testing RMSE   \\
  \midrule
\multirow{2}{*}{Airfoil Self-Noise}&BLS& $(N_f, N_w, N_e)\!=\!(1, 5, 166)$  & 0.0094 $\!\pm\!$ 0.0012 & 0.0651 $\!\pm\!$ 0.0045 & 0.0959 $\!\pm\!$ 0.0137 \\
 &C-BLS &  $(N_f, N_w, N_e, \sigma)\!=\!(1, 20, 136, 2^{\!-\!2}) $ &  0.1099 $\!\pm\!$ 0.0047 & 0.0674 $\!\pm\!$ 0.0022 & \textbf{0.0923 $\!\pm\!$ 0.0074} \\
 \midrule
  \multirow{2}{*}{Bodyfat}& BLS & $(N_f, N_w, N_e)\!=\!(13, 17, 1)$  & 0.0151 $\!\pm\!$ 0.0019 & 0.0080 $\!\pm\!$ 0.0001 & 0.0054 $\!\pm\!$ 0.0001 \\
 &C-BLS &  $(N_f, N_w, N_e, \sigma)\!=\!(3, 12, 21, 2^{\!-\!5}) $ &  0.0161 $\!\pm\!$ 0.0047 & 0.0073 $\!\pm\!$ 0.0007 & \textbf{0.0040 $\!\pm\!$ 0.0007}\\
 \midrule
 \multirow{2}{*}{Cleveland}&BLS& $(N_f, N_w, N_e)\!=\!(1, 1, 151)$  & 0.0020 $\!\pm\!$ 0.0004 & 0.1106 $\!\pm\!$ 0.0011 & 0.1312 $\!\pm\!$ 0.0100 \\
 &C-BLS &  $(N_f, N_w, N_e, \sigma)\!=\!(1, 1, 21, 2^{\!-\!1}) $ &  0.0179 $\!\pm\!$ 0.0017 & 0.1104 $\!\pm\!$ 0.0013 & \textbf{0.1306 $\!\pm\!$ 0.0076} \\
  \midrule
  \multirow{2}{*}{Forestfires}& BLS & $(N_f, N_w, N_e)\!=\!(1, 1, 1)$  & 0.0010 $\!\pm\!$ 0.0001 & 0.0593 $\!\pm\!$ 0.0001 & \textbf{0.0561 $\!\pm\!$ 0.0001} \\
 &C-BLS &  $(N_f, N_w, N_e, \sigma)\!=\!(1, 1, 1, 2^{\!-\!3}) $ &  0.0020$\!\pm\!$ 0.0002 & 0.0595 $\!\pm\!$ 0.0000 & 0.0563 $\!\pm\!$ 0.0001\\
  \midrule
 \multirow{2}{*}{Mortgage}&BLS& $(N_f, N_w, N_e)\!=\!(3, 9, 141)$  & 0.0107 $\!\pm\!$ 0.0010 & 0.0030 $\!\pm\!$ 0.0001 & \textbf{0.0056 $\!\pm\!$ 0.0003} \\
&C-BLS &  $(N_f, N_w, N_e, \sigma)\!=\!(3, 9, 141, 2^{\!-\!4}) $ &  0.0784 $\!\pm\!$ 0.0029 & 0.0030 $\!\pm\!$ 0.0001 & \textbf{0.0056 $\!\pm\!$ 0.0003}\\
 \midrule
 \multirow{2}{*}{Quake}& BLS & $(N_f, N_w, N_e)\!=\!(19, 19, 1)$  & 0.0378 $\!\pm\!$ 0.0023 & 0.1711 $\!\pm\!$ 0.0002 & \textbf{0.1728 $\!\pm\!$ 0.0003} \\
 & C-BLS &  $(N_f, N_w, N_e, \sigma)\!=\!(19, 19, 1, 2^{1}) $ &  0.2364 $\!\pm\!$ 0.1275 & 0.1711 $\!\pm\!$ 0.0002 & \textbf{0.1728 $\!\pm\!$ 0.0003}\\
  \midrule
 \multirow{2}{*}{Weather Izmir}&BLS& $(N_f, N_w, N_e)\!=\!(3, 9, 31)$  & 0.0081 $\!\pm\!$ 0.0006 & 0.0175 $\!\pm\!$ 0.0003 & 0.0206 $\!\pm\!$ 0.0004 \\
 &C-BLS &  $(N_f, N_w, N_e, \sigma)\!=\!(1, 16, 56, 2^{\!-\!5}) $ &  0.0548 $\!\pm\!$ 0.0026 & 0.0173 $\!\pm\!$ 0.0011 & \textbf{0.0200 $\!\pm\!$ 0.0005}\\
 \bottomrule
 \end{tabular}
 \label{Tab2}
  \end{center}
\end{table*}
\begin{table*}[htbp]
 \renewcommand\arraystretch{1.2}
\newcommand{\tabincell}[2]{\begin{tabular}{@{}#1@{}}#2\end{tabular}}
 \caption{Performance comparison of BLS and C-BLS on classification benchmark datasets}
 \begin{center}
 \begin{tabular}{llccccccc}
  \toprule
Dataset & Algorithm  & Parameters & Training Time (s)    & Training Accuracy  & Testing Accuracy   \\
  \midrule
\multirow{2}{*}{Balance Scale}&BLS& $(N_f, N_w, N_e)\!=\!(1, 11, 31)$  & 0.0079 $\!\pm\!$ 0.0004 & 91.34 $\!\pm\!$ 0.38 & \textbf{90.70 $\!\pm\!$ 1.00} \\
 & C-BLS &  $(N_f, N_w, N_e, \sigma)\!=\!(1, 18, 41, 2^{4}) $ &  0.0401 $\!\pm\!$ 0.0027 & 91.94 $\!\pm\!$ 0.51 & 90.63 $\!\pm\!$ 1.10\\
  \midrule
\multirow{2}{*}{Chess}&BLS& $(N_f, N_w, N_e)\!=\!(3, 12, 176)$  & 0.0255 $\!\pm\!$ 0.0015 & 97.10 $\!\pm\!$ 0.43  & \textbf{95.44 $\!\pm\!$ 0.59} \\
 &C-BLS &  $(N_f, N_w, N_e, \sigma)\!=\!(3, 12, 176, 2^{5}) $ &  0.0452 $\!\pm\!$ 0.0040 & 97.10 $\!\pm\!$ 0.43 & \textbf{95.44 $\!\pm\!$ 0.59} \\
  \midrule
  \multirow{2}{*}{Diabetes}&BLS& $(N_f, N_w, N_e)\!=\!(3, 12, 6)$  & 0.0090 $\!\pm\!$ 0.0004 & 79.04 $\!\pm\!$ 0.66 & 75.68 $\!\pm\!$ 1.03 \\
 & C-BLS &  $(N_f, N_w, N_e, \sigma)\!=\!(7, 16, 6, 2^{1}) $ &  0.0566 $\!\pm\!$ 0.0027 & 79.27 $\!\pm\!$ 0.50 & \textbf{75.92 $\!\pm\!$ 1.02} \\
  \midrule
  \multirow{2}{*}{Ecoli}&BLS& $(N_f, N_w, N_e)\!=\!(1, 9, 11)$  & 0.0063 $\!\pm\!$ 0.0003 & 89.91 $\!\pm\!$ 0.68 & \textbf{86.18 $\!\pm\!$ 2.03} \\
 &C-BLS &  $(N_f, N_w, N_e, \sigma)\!=\!(1, 9, 11, 2^{5}) $ &  0.0221 $\!\pm\!$ 0.0048 & 89.91 $\!\pm\!$ 0.68 & \textbf{86.18 $\!\pm\!$ 2.03} \\
  \midrule
  \multirow{2}{*}{Horse Colic}&BLS& $(N_f, N_w, N_e)\!=\!(17, 17, 21)$  & 0.0166 $\!\pm\!$ 0.0010 & 89.67 $\!\pm\!$ 0.87 & 80.81 $\!\pm\!$ 2.68 \\
 & C-BLS &  $(N_f, N_w, N_e, \sigma)\!=\!(11, 13, 1, 2^{4}) $ &  0.0501 $\!\pm\!$ 0.0025 & 88.87 $\!\pm\!$ 0.23 & \textbf{82.21 $\!\pm\!$ 0.66} \\
  \midrule
  \multirow{2}{*}{Wireless}&BLS& $(N_e, N_f, N_e)\!=\!(5, 8, 161)$  & 0.0157 $\!\pm\!$ 0.0010 & 98.46 $\!\pm\!$ 0.18 & 97.53 $\!\pm\!$ 0.44 \\
 & C-BLS &  $(N_e, N_f, N_e, \sigma)\!=\!(5, 8, 161, 2^{3}) $ &  0.1696 $\!\pm\!$ 0.0044 & 98.43 $\!\pm\!$ 0.17 & \textbf{97.56 $\!\pm\!$ 0.44} \\
 \bottomrule
 \end{tabular}
 \label{Tab3}
  \end{center}
\end{table*}

\subsubsection{Comparison with Robust BLS Variants}
Although C-BLS, in general, performs better than the standard BLS in the aforementioned thirteen data sets in terms of testing RMSE or testing accuracy, it is hard to say that C-BLS is robust to outliers, since we don't know any prior information about the noise distribution of these real datasets. To test the robustness of C-BLS, we choose the dataset ``Bodyfat" and the dataset ``Chess" as two representative examples. In detail, for dataset ``Bodyfat",
the targets of training samples are added with random outliers from the interval $[0, 1]$, with the
contamination level of $p$; for dataset ``Chess", a part of training samples (with the ratio of $p$) are randomly selected and their class labels are reversed to its opposite side.
Moreover, we note that there are several robust BLS variants that are also realized by replacing the MMSE criterion with other robust one, i. e, the robust BLS with $l_{2}$-norm regularization (L2RBLS) \cite{BLS_L1} and the weighted BLS (WBLS) \cite{WBLS}. Therefore, not only the standard BLS but also these two methods are chosen as the benchmarks for comparison.
To make all comparative algorithms reach their best performance as much as possible, we also perform a grid search for their parameters.
In detail, all the aforementioned parameters search ranges of the standard BLS are shared with L2RBLS since they share the same architecture with each other.
For WBLS, the Huber weight function with cut-value $b$ being chosen from $\{0.1, 0.3, 0.5, 0.7, 1, 3, 5, 7, 10\}$ is adopted as the weighted penalty factor and the regularization parameter is fixed at $2^{-30}$. Then, we search its optimal number of enhancement nodes from $[1, 600]$ with the step-size of $10$.
\begin{figure}[htbp]
\centering
\subfigure[Regression dataset: Bodyfat]{
\label{fig2a} 
\includegraphics[width=3.0in]{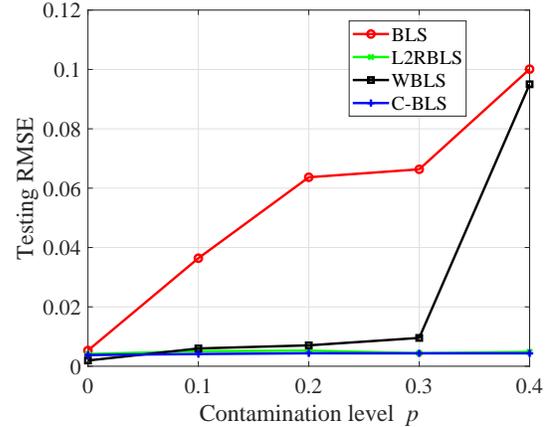}}
\hspace{0.05in}
\subfigure[Classification dataset: Chess]{
\label{fig2b} 
\includegraphics[width=3.0in]{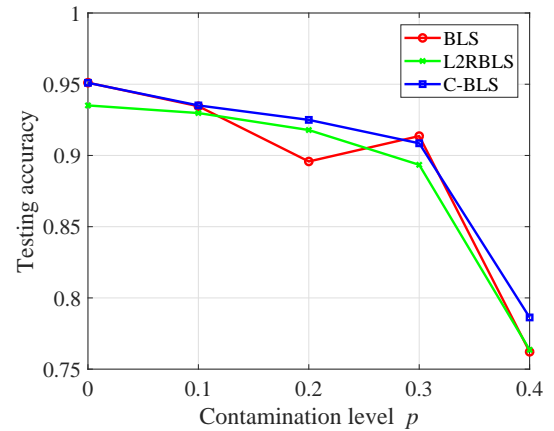}}
\caption{Performance comparison of different algorithms under different contamination level $p$.}
\label{fig2}
\end{figure}

Fig.~\ref{fig2} shows the comparative results of different methods versus contamination level $p$.
It can be seen from Fig.~\ref{fig2a} that, with the increment of $p$, the testing RMSE of BLS increases evidently while the testing RMSEs of L2RBLS, WBLS and C-BLS are not very sensitive to $p$. Moreover, the proposed C-BLS can, in general, achieve a smaller testing RMSE compared with L2RBLS and WBLS. Similarly, we can see from Fig.~\ref{fig2b} that, the proposed C-BLS also performs better than others when applied to classification tasks.
It is worth noting that the weighted operators provided in \cite{WBLS} are originally designed for multiple-input-single-output regression cases, so we don't apply the corresponding WBLS to classification datasets in this part. Moreover, the robustness of C-BLS seems to be not as strong as those in regression examples, since its testing accuracy decreases evidently like the standard BLS and L2RBLS when $p=0.4$. A possible reason is that the error distribution in classification applications is rather complicated compared to these multiple-input-single-output regression examples. 
\subsubsection{Convergence Curves}
\begin{figure}[htbp]
\centering
\subfigure[Regression datasets]{
\label{fig3a} 
\includegraphics[width=3.0in]{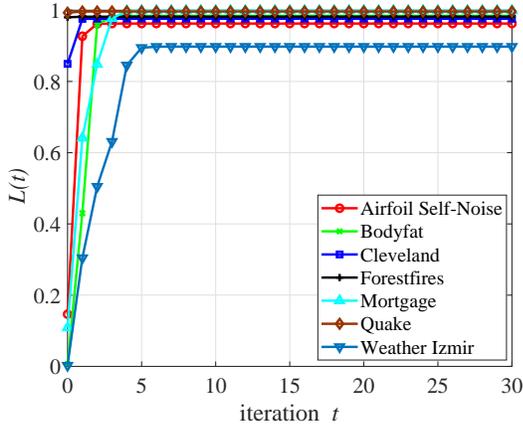}}
\hspace{0.05in}
\subfigure[Classification datasets]{
\label{fig3b} 
\includegraphics[width=3.0in]{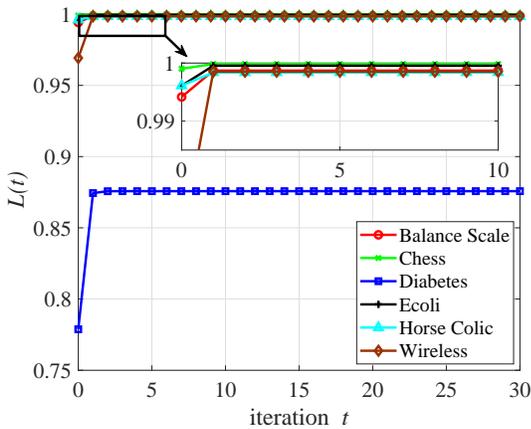}}
\caption{Convergence curves of C-BLS on benchmark datasets.}
\label{fig3}
\end{figure}

We now investigate the convergence of the proposed C-BLS. For this purpose, the value of the normalized objective function at iteration $t$ is calculated by
\begin{equation}\label{E1000}
\resizebox{0.88\hsize}{!}{$
L(t)\!=\!\frac{1}{N}\left(\sum_{i \!=\! 1}^N {\exp ( \!-\! \frac{{{{\left\| {{\textbf{u}_i} \textbf{W}(t)  \!-\! {\textbf{y}_i}} \right\|}_{2}^2}}}{{2{\sigma ^2}}})}\!-\!\frac{\lambda}{2}\parallel\textbf{W}(t)\parallel_{2}^{2}\right).
$}
\end{equation}
If the value of $L(t)$ is not decreasing, the convergence of the proposed C-BLS will be guaranteed.
Fig.~\ref{fig3} shows the variation of the normalized objective function $L(t)$ versus iteration $t$.
Herein, all parameters are set as the same of those in Tables~\ref{Tab2} and~\ref{Tab3}.
It can be seen from Fig~\ref{fig3a} and Fig~\ref{fig3b} that, the value of $L(t)$ increases at the beginning and finally reach a stable value. This conforms the convergence of the proposed C-BLS in our experiments. In addition, the stable value of C-BLS can be reached within $10$ iterations for most datasets. Thus, the number of maximum iteration may be not necessary set to be a very large value for practical applications.

\subsubsection{Parameters Sensitivity}
\begin{figure}[htbp]
\centering
\subfigure[Regression datasets]{
\label{fig4a} 
\includegraphics[width=3.0in]{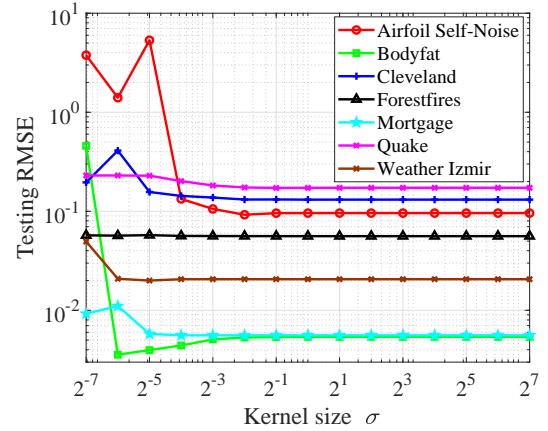}}
\hspace{0.05in}
\subfigure[Classification datasets]{
\label{fig4b} 
\includegraphics[width=3.0in]{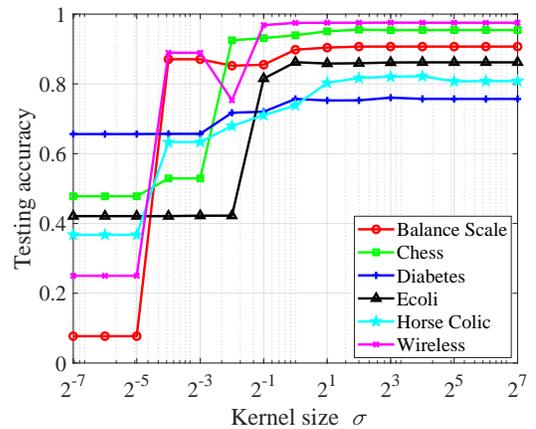}}
\caption{The influence of kernel size $\sigma$ on learning performance.}
\label{fig4}
\end{figure}

Compared with the standard BLS, an additional parameter, i. e., kernel size $\sigma$, is introduced in C-BLS. In this subsection, the influence of $\sigma$ on the learning performance of C-BLS is investigated.
Therefore, we change the value of $\sigma$ in a candidate set $\{2^{-7}, 2^{-6}, \cdots, 2^6, 2^7\}$, while other parameters are determined in the same way detailed before. Fig.~\ref{fig4a} and Fig.~\ref{fig4b}, respectively, show the testing RMSEs and testing accuracies on regression and classification datasets. As can be seen from Fig.~\ref{fig4a} and Fig.~\ref{fig4b}, although the influence of $\sigma$ on learning performance varies with datasets, there is always a range of $\sigma$ in which the C-BLS performs better than others when a specific dataset is considered.
Hence, $\sigma$ is a parameter that should be appropriately chosen for C-BLS. Except for the grid search method adopted in this work, other productive technologies, like Silverman rule \cite{MCC-PCA, Silverman}, can also be good candidates to determine $\sigma$ in practice.

\subsection{Performance Evaluation of the Incremental Learning Algorithms}
For large-scale datasets, there are usually a large number of samples and a high dimension.
Once some new samples arrive or the network deems to be expanded, C-BLS needs to
run a complete training cycle for remodeling the learning system, which not only results in large time consumption, but also requires large storage resources.
As mentioned before, to guarantee that the system can be remodeled quickly without the entire retraining process from the beginning when some new samples arrive or the network deems to be expanded, we also develop several alternative incremental learning algorithms.
In this part, we present experiments to demonstrate their effectiveness.
Similar to \cite{BLS2018}, the Mixed National Institute of Standards
and Technology (MINST) handwriting dataset \cite{MNIST} which contains $60000$ training samples and $10000$ testing samples is adopted here for study purpose, and one can refer to \cite{BLS2018, MNIST} for details about this dataset.
\begin{table}[htbp]
 \renewcommand\arraystretch{1.2}
\newcommand{\tabincell}[2]{\begin{tabular}{@{}#1@{}}#2\end{tabular}}
 \caption{Recognition results on MNIST dataset at different incremental steps: increment of training samples.}
 \begin{center}
 \begin{tabular}{cccc}
  \toprule
 Training Samples    & Training Time (s)   & Testing Accuracy    \\
  \midrule
  $20000$  & 104.02 $\!\pm\!$ 2.58  & 97.92 $\!\pm\!$ 0.05 \\
  $20000\rightarrow30000$  & 50.58 $\!\pm\!$ 1.77  &  98.12 $\!\pm\!$ 0.04\\
  $30000\rightarrow40000$  & 50.94 $\!\pm\!$ 1.68  &  98.19 $\!\pm\!$ 0.05\\
  $40000\rightarrow50000$  & 51.17 $\!\pm\!$ 2.27  &  98.26 $\!\pm\!$ 0.04\\
  $50000\rightarrow60000$  & 51.54 $\!\pm\!$ 0.96  &  98.30 $\!\pm\!$ 0.05 \\
 \bottomrule
 \end{tabular}
 \label{Tab4}
  \end{center}
\end{table}

First, the receiving of new training samples is considered. Hence, we use the first $20000$ samples to train the initial network, and then add $10000$ additional samples to the network each time until all training samples are utilized. During the whole process, the $N_f$, $N_m$, $N_e$, $\gamma$ and $\sigma$ are respectively fixed at $10$, $10$, $5000$, $10^{-5}$ and $2^{5}$, respectively.
Table~\ref{Tab4} shows the training times and testing accuracies at different incremental steps when the network receives the new training samples. It can be seen from Table~\ref{Tab4} that, with the increment of the training samples, the testing accuracy increases as expected, and finally accuracy  reaches at $98.30\pm0.05$, which is comparable to the standard BLS when the same network allocation is adopted (see Table V of Ref.~\cite{BLS2018} for details).
Meanwhile, other than the initial stage, the subsequent incremental learning steps don't need to consume too much training time. This conforms the effective of the proposed incremental learning algorithms when some new training samples are received.
%
\begin{table}[htbp]\smaller
 \renewcommand\arraystretch{1.2}
\newcommand{\tabincell}[2]{\begin{tabular}{@{}#1@{}}#2\end{tabular}}
 \caption{Recognition results on MNIST dataset at different incremental steps: increments of feature nodes and enhancement nodes.}
 \begin{center}
 \begin{tabular}{ccc}
  \toprule
  (Feature Nodes, Enhancement Nodes) & Training Time (s) &  Testing Accuracy      \\
  \midrule
  ($10\times6,3000$) & 108.90 $\!\pm\!$ 1.60 &  97.96 $\!\pm\!$ 0.09 \\
  ($10\times6\!\rightarrow\!10\times7$, $3000\!\rightarrow\!3500$)  & 15.02 $\!\pm\!$ 0.61  &  98.06 $\!\pm\!$ 0.14\\
  ($10\times7\!\rightarrow\!10\times8$, $3500\!\rightarrow\!4000$)  & 16.98 $\!\pm\!$ 0.51  &  98.16 $\!\pm\!$ 0.09 \\
  ($10\times8\!\rightarrow\!10\times9$, $4000\!\rightarrow\!4500$)  & 22.36 $\!\pm\!$ 1.11  &  98.25 $\!\pm\!$ 0.09\\
  ($10\times9\!\rightarrow\!10\times10$, $4500\!\rightarrow\!5000$) & 25.03 $\!\pm\!$ 0.99  &  98.29 $\!\pm\!$ 0.09\\
 \bottomrule
 \end{tabular}
 \label{Tab5}
  \end{center}
\end{table}

Then, the increments of feature nodes and enhancement nodes are considered. The training samples in this case are received in one time. For study purpose, we firstly train the initial network with $10\times6$ feature nodes and $3000$ enhancement nodes. Subsequently, the additional feature nodes and enhancement nodes are added to the network for performance improvement. In detail, the number of additional feature nodes in each incremental step is fixed at $10\times1$, while the number of additional enhancement nodes in each incremental step is set as $500$ (the first $200$ correspond to the additional feature nodes, and the rest $300$ are extra inserted).
In Table~\ref{Tab5}, we show the training times and testing accuracies at different incremental steps when both new feature nodes and enhancement nodes are inserted to network.
It can be seen from Table~\ref{Tab5} that, with the numbers of both feature nodes and enhancement nodes increase, the algorithm has the chance to get a higher recognition accuracy than the initial stage, while the training time in each incremental step don't need to be very high.
\begin{table}[htbp]\smaller
 \renewcommand\arraystretch{1.2}
\newcommand{\tabincell}[2]{\begin{tabular}{@{}#1@{}}#2\end{tabular}}
 \caption{Recognition results on MNIST dataset at different incremental steps: increments of training samples and enhancement nodes.}
 \begin{center}
 \begin{tabular}{ccc}
  \toprule
 (Training Samples, Enhancement Nodes)  & Training Time(s) & Testing Accuracy    \\
  \midrule
  ($20000,1000$) & 8.44 $\!\pm\!$ 0.22 &  97.02 $\!\pm\!$ 0.12 \\
  ($20000\!\rightarrow\!30000$, $1000\!\rightarrow\!2000$)  & 31.45 $\!\pm\!$ 0.97  &  97.66 $\!\pm\!$ 0.06\\
  ($30000\!\rightarrow\!40000$, $2000\!\rightarrow\!3000$)  & 42.35 $\!\pm\!$ 0.70  &  98.04 $\!\pm\!$ 0.06\\
  ($40000\!\rightarrow\!50000$, $3000\!\rightarrow\!4000$)  & 57.94 $\!\pm\!$ 1.52  &  98.08 $\!\pm\!$ 0.41\\
  ($50000\!\rightarrow\!60000$, $4000\!\rightarrow\!5000$)  & 78.87 $\!\pm\!$ 2.25  &  98.39 $\!\pm\!$ 0.07\\
 \bottomrule
 \end{tabular}
 \label{Tab6}
  \end{center}
\end{table}

Finally, the synchronous increments of training samples and enhancement nodes are considered. At the initial stage, only the first $20000$ training samples are sent to network, while the number of enhancement nodes is set as $1000$.
At each subsequent incremental step, $10000$ new training samples are sent to network and the additional number of enhancement nodes is set as $1000$. Table~\ref{Tab6} shows the training times and testing accuracies at different incremental steps. As can be seen from Table~\ref{Tab6}, the proposed incremental learning algorithms can work well when both training samples and enhancement nodes increase.

\subsection{Application to Time Series Prediction}
In this subsection, the proposed C-BLS will be applied to perform time series prediction tasks.
The presented examples include Mackey-Glass time series prediction and monthly mean total
``Sunspot Number" prediction.


\subsubsection{Data sets}
Mackey-Glass time series
displays the characteristics of chaotic dynamics, and can be
described by the following time-delay differential equation:
\begin{equation}\label{EE1}
  \frac{d x(t)}{dt}=-b x(t)+\frac{a x(t-\tau)}{1+x(t-\tau)^{10}}
\end{equation}
where $a=0.1$, $b=0.2$ and $\tau=30$ are the default choice in the current paper. Based on \eqref{EE1}, we first generate $1200$ data points, among which, the first $1000$ are used for training and the rest $200$ are used for testing.
To comprehensively evaluate the performance of the proposed methods, the Gaussian signal sequence and the impulse signal sequence are respectively added to the training samples to model the influences caused by different disturbances. In detail, the Gaussian signal sequence is generated by a Gaussian distribution with zero-mean and variance of $0.01$; the impulse signal sequence is generated by an $\alpha$-stable distribution \cite{KRMC, QKMC} whose characteristic function can be simply formulated by $\psi(\omega)={\rm exp}(-\gamma|\omega|^{\alpha})$, where $\gamma=0.1$ and $\alpha=1.5$. As suggested in \cite{QKMC}, we adopt the previous seven points to predict the current one. Hence, the input and output at time $t$ have the form of $\textbf{x}_{t}=[x(t-7),\cdots,x(t-1)]$ and $y_{t}=x(t)$, respectively.

The ``Sunspot Number" is a real time series dataset which records the number of monthly mean total sunspot from January $1749$ to December $2017$, and is available at http://www.sidc.be/silso/datafiles. In the following experiments, the data are firstly normalized into the range of $[0,1]$. After that, the training set is constructed based on the sunspot number recorded from January $1749$ to December $1990$, while the testing set is constructed based on the sunspot number recorded from January $1991$ to December $2017$.
Like \cite{KAF}, the embedding dimension and the delay time are respectively set as $4$ and $1$ in the experiments.

\subsubsection{Experimental Results}
When faced with time series prediction tasks, the standard BLS \cite{BLS_U}, the structured manifold BLS (SM-BLS) \cite{BLS_SM}, the robust manifold BLS (RM-BLS) \cite{RMBLS} and the recurrent BLS \cite{BLS_R} have all been proven to be good candidates. Moreover, we note that the first three of them are developed based on the fully feedforward neural network model like the proposed C-BLS, so they are chosen as the benchmarks for comparison.

Table~\ref{Tab7} shows the main parameters setting and comparative results of different algorithms. It can be seen from Table~\ref{Tab7} that, if the noise sequence is drawn from Gaussian distribution, the standard BLS and SMBLS perform better than that of RMBLS and C-BLS. However, if the Gaussian assumption is not satisfied, their performance can be degenerated evidently. In contrary, the C-BLS can always obtain a relatively satisfactory prediction result in both Gaussian and alpha-stable noise environments.
Moreover, the prediction results on real Sunspot number dataset also conform that C-BLS is a good candidate for time series prediction tasks.
\begin{table*}[htbp]
 \renewcommand\arraystretch{1.1}
\newcommand{\tabincell}[2]{\begin{tabular}{@{}#1@{}}#2\end{tabular}}
 \caption{Performance comparison of different methods on time series datasets.}
 \begin{center}
 \begin{tabular}{llcccccc}
  \toprule
Dataset & Algorithm  & Parameters & Training Time (s)    & Testing RMSE   \\
  \midrule
\multirow{2}{*}{Mackey-Glass (Gaussian)}&BLS& $(N_f, N_w, N_e, \gamma)\!=\!(7, 13, 160, 10^{\!-\!5})$  & 0.0294 $\!\pm\!$ 0.0111 & 0.0483 $\!\pm\!$ 0.0042 \\
 &SM-BLS &  $(N_f, N_w, N_e, \lambda, \alpha, \beta)\!=\!(15, 1, 180, 10^{\!-\!4}, 10^{\!-\!5}, 10^{\!-\!2}) $ &  0.7158 $\!\pm\!$ 0.0158  & \textbf{0.0477 $\!\pm\!$ 0.0035}\\
  &RM-BLS &  $(N_f, N_w, N_e, \lambda, \alpha, \beta)\!=\!(27, 1, 430, 10^{\!-\!2}, 10^{0}, 10^{\!-\!1}) $ &  0.2902 $\!\pm\!$ 0.0071 &  0.0484 $\!\pm\!$ 0.0082\\
   &C-BLS &  $(N_f, N_w, N_e, \gamma, \sigma)\!=\!(7, 15, 160, 10^{\!-\!4}, 2^{5}) $ &  0.0487 $\!\pm\!$ 0.0030 &  0.0486 $\!\pm\!$ 0.0051\\
 \midrule
  \multirow{2}{*}{Mackey-Glass ($\alpha$-stable)}& BLS & $(N_f, N_w, N_e, \gamma)\!=\!(25, 17, 270, 10^{\!-\!4})$  & 0.0631 $\!\pm\!$ 0.0041  & 0.0909 $\!\pm\!$ 0.0123 \\
 &SM-BLS &  $(N_f, N_w, N_e, \lambda, \alpha, \beta)\!=\!(29, 1, 420, 10^{\!-\!7}, 10^{0}, 10^{\!-\!2}) $ &  1.0977 $\!\pm\!$ 0.0196 &  0.0921 $\!\pm\!$ 0.0030\\
  &RM-BLS &  $(N_f, N_w, N_e, \lambda, \alpha, \beta)\!=\!(27, 1, 360, 10^{\!-\!7}, 10^{\!-\!1}, 10^{\!-\!2}) $ &  0.2834 $\!\pm\!$ 0.0053 &  0.0873 $\!\pm\!$ 0.0045\\
  &C-BLS &  $(N_f, N_w, N_e, \gamma, \sigma)\!=\!(25, 20, 456, 10^{\!-\!4}, 2^{\!-\!2}) $ &  1.6315 $\!\pm\!$ 0.0283 &  \textbf{0.0547 $\!\pm\!$ 0.0035}\\
  \midrule
  \multirow{2}{*}{Sunspot Number}& BLS & $(N_f, N_w, N_e, \gamma)\!=\!(25, 13, 150, 10^{\!-\!3})$  & 0.0697 $\!\pm\!$ 0.0019  & 0.0543 $\!\pm\!$ 0.0013 \\
 &SM-BLS &  $(N_f, N_w, N_e, \lambda, \alpha, \beta)\!=\!(23, 1, 50, 10^{\!-\!5}, 10^{\!-\!2}, 10^{\!-\!2}) $ &  6.0065 $\!\pm\!$ 0.1869 &  0.0536 $\!\pm\!$ 0.0004\\
  &RM-BLS &  $(N_f, N_w, N_e, \lambda, \alpha, \beta)\!=\!(7, 1, 90, 10^{\!-\!2}, 10^{\!-\!1}, 10^{\!-\!3}) $ &  1.3597 $\!\pm\!$ 0.0118 &  0.0531 $\!\pm\!$ 0.0006\\
   &C-BLS &  $(N_f, N_w, N_e, \gamma, \sigma)\!=\!(29, 13, 30, 10^{\!-\!2}, 2^{\!-\!3}) $ &  0.4470 $\!\pm\!$ 0.0913  & \textbf{0.0530 $\!\pm\!$ 0.0003}\\
 \bottomrule
 \end{tabular}
 \label{Tab7}
  \end{center}
\end{table*}

\subsection{Application to EEG Classification}
\begin{figure}[htbp]
\centering
\includegraphics[width=3.1in]{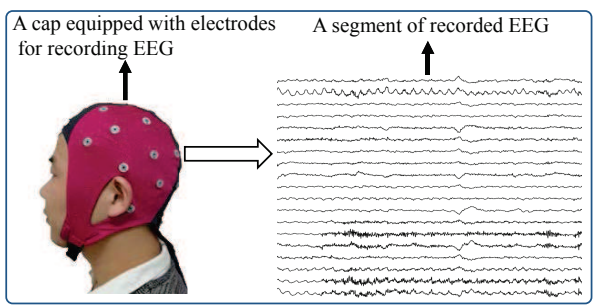}
\caption{A glimpse of recording EEG.}
\label{fig6}
\end{figure}

Electroencephalography (EEG) is a kind of neurophysiological signal, which is generated by the firing of neurons in the brain and usually collected via electrodes placed on the scalp surface (see Fig.~\ref{fig6} for a glimpse).
Due to its safety, inexpensiveness, non-invasiveness and a favorable
temporal resolution (on the precision of milliseconds), EEG has been becoming an indispensable role in the applications of Brain-Compute-Interface (BCI) \cite{BCI}.
During the past few years, many efforts have been devoted to improving the EEG classification accuracy of the EEG-based BCI systems. However, because of the inherent complexity of EEG signals (such as high dimensionality, high dynamic and low signal-noise-ratio (SNR)), the corresponding classification result, up to today, still cannot meet the application requirements.
In this subsection, we will try to apply the proposed C-BLS to EEG classification tasks.

\subsubsection{Data sets}
The benchmark dataset adopted here is the data set 2b of BCI competion IV \cite{EEG2b}, which includes the EEG recordings from nine healthy subjects. For each subject, there are five sessions about two-classes motor imagery tasks (left hand VS right hand). Among the five sessions, the first two of them were recorded without feedback and each of them includes $120$ trials, while the other three sessions were incorporated with online feedback and each of them has $160$ trials. In each session, three channels including C3, Cz and C4 were used to record EEG measurements with a sampling frequency of $250$ HZ. To remove the influence caused by the power equipments, the EEG sequences had been bandpass filtered from $0.5$ HZ to $100$ HZ and a notch filter at the notch frequency of $50$ HZ had also been adopted. In the following experiments, the EEG recordings from the first three sessions are used for training and the EEG recordings from the last two sessions are used for testing.
\subsubsection{Reprocessing and Feature Extraction}
In order to exclude the EEG recordings that are not related to motor imagery tasks, only the segment between $0.5$ s and $2.5$ s after cue is kept in each trial. These reserved EEG segments are subsequently bandpass filtered from $8$ HZ to $35$ HZ as recommended in \cite{EEG2b}. Moreover, since the brain activities reflecting the imaginary limb motion mainly exist in the contralateral region of the
brain, the data from channel Cz is removed in our experiments.
Without loss of generality,
the MATLAB command ``pwelch", with a sliding Hamming
window of $0.25$ s and an overlap of $0.125$ s, is used for computing the power spectrum
density (PSD) \cite{Pwelch, PSD_REF} of the EEG data in C3 and C4. Similar to \cite{PSD_REF}, the number of discrete-Fourier-transform (DFT) points is set as $256$, and the final feature vector for each trail is obtained by concatenating the PSD values
of C3 and C4 channels.

\subsubsection{Parameters Setting and Results}
Other than the standard BLS, several classical classifiers, including K-nearest neighbors (K-NN) \cite{KNN}, support vector machine (SVM) \cite{SVM}, extreme learning machine (ELM) \cite{ELM-T},
DBN \cite{DBN} and SAE \cite{SAE}, are also chosen as the benchmark for comparison. The parameter search ranges of different algorithms are as follows.
\begin{itemize}
  \item For BLS and C-BLS, the regularization parameter in them are both chosen from the range of $\{10^{-5}, 10^{-4},\cdots,10^{4}, 10^{5}\}$, while the remainder parameter are searched within the ranges detailed in Section~\ref{Sec4A}.
  \item For K-NN, we choose the number of $k$ from $1$ to $10$ with the step-size of $1$.
  \item For SVM, the Gaussian kernel is adopted. After that, the kernel size and regularization parameter are searched from $\{2^{-5}, 2^{-4},\cdots,2^{4}, 2^{5}\}$ and  $\{10^{-5}, 10^{-4},\cdots,10^{4}, 10^{5}\}$, respectively.
  \item For ELM, the sigmoid activation function is adopted with the input weights and biases randomly generated from $[-1, 1]$ and $[0, 1]$, respectively. Meanwhile, we perform a grid search for its number of hidden nodes from $[5:600]$ with the step size of $5$ and regularization parameter from $\{10^{-5}, 10^{-4},\cdots,10^{4}, 10^{5}\}$.
  \item For DBN and SAE, they are implemented using ``DeepLearnToolbox" \cite{DLT}.
Herein, only two hidden-layers are involved, and the optimal number of neurons at the first and the second layer are both chosen
from $[20:300]$ with the step-size of $20$. In addition, the number of epoches for unsupervised and supervised learning stages are respectively set as $150$ and $100$, the learning rates for unsupervised and supervised learning stages are respectively set as $1.5$ and $1.0$, the mini-batch sizes for supervised and unsupervised learning stages are both fixed at $20$, and the activation function in the network is set as the sigmoid function.
\end{itemize}


\begin{table}[htbp] \smaller
 \renewcommand\arraystretch{1.3}
 \caption{Testing accuracies of different classifiers on data set IIb of BCI competition IV (in \%) }
 \begin{center}
 \begin{tabular}{lccccccc}
  \toprule
\multirow{2}{*}{Subject} & \multicolumn{7}{c}{Classifier} \\
\cline{2-8}
   & K-NN & SVM & ELM & SAE & DBN & BLS & C-BLS \\
  \midrule
  B01  & 60.31 & 69.69 & 67.34 & 67.69 & 65.44 & 71.03 & \textbf{71.22}\\
  B02  & 56.43 & 61.79 & 59.57 & 57.18 & 54.82 & 60.21 & \textbf{61.89}\\
  B03  & 53.44 & \textbf{54.38} & 53.47 & 52.84 & 52.37 & 52.97 & 52.97  \\
  B04  & \textbf{96.56} & 95.31 & 95.91 & 95.56 & 93.09 & 93.91 & 93.50  \\
  B05  & 70.94 & \textbf{87.81} & 87.28 & 86.59 & 83.88 & 86.34 & 87.25  \\
  B06  & 75.00 & 78.44 & 79.66 & 78.59 & 71.31 & 79.97 & \textbf{82.22} \\
  B07  & 67.81 & 64.69 & 70.50 & 70.63 & 63.88 & 71.81 & \textbf{71.97} \\
  B08  & 88.44 & 90.00 & 91.25 & 90.22 & 85.03 & 90.28 & \textbf{91.59} \\
  B09  & 78.75 & 80.31 & 80.12 & 79.78 & 79.09 & 80.38 & \textbf{80.59}\\
  Mean & 71.96 & 75.82 & 76.12 & 75.45 & 72.10 & 76.32 & \textbf{77.02} \\
 \bottomrule
 \end{tabular}
 \label{Tab8}
 \end{center}
\end{table}

Table~\ref{Tab8} shows the classification results of different methods. It can be seen from Table~\ref{Tab8} that, C-BLS can obtain a higher testing classification accuracy on most subjects and it also has the highest average classification accuracy on all subjects. These results suggest that the proposed C-BLS can be, at least, a better candidate compared with the standard BLS and several other classical classifiers when applied to EEG classification tasks.

\section{Conclusion} \label{Sec5}
To enhance the robustness of Broad Learning System (BLS), the maximum correntropy criterion (MCC) was introduced to train its output weights, generating a correntropy based BLS (C-BLS).
To ensure that the system can be updated quickly without the entire retraining process from the beginning when some new samples arrive or the network deems to be expanded, three alternative incremental learning algorithms were then developed.
As we know, this is the first attempt to combine the Information Theoretic Learning (ITL) criterion and the BLS architecture.
Experimental results showed that the proposed C-BLS is not only robust to outliers but also can be a good candidate if we don't know any prior information about the noise distribution in some practical scenarios. Meanwhile, the proposed incremental learning algorithms were proven to have the ability to support rapid remodeling.
In the future, the flexibly combinations of C-BLS with some well-developed feature mapping technologies, like convolution-pooling operation \cite{BLS_U}, neuro-fuzzy model \cite{BLS_F}, and structured manifold learning method \cite{BLS_SM}, will be considered for further performance improvement.




%

\appendices


\ifCLASSOPTIONcaptionsoff
  \newpage
\fi



\bibliographystyle{IEEEtran}
\bibliography{References}
\end{document}